\pdfoutput=1
\documentclass[review]{elsarticle}
\usepackage{subfigure}
\usepackage{longtable}
\usepackage{algorithm, algorithmic}
\usepackage{bm}
\usepackage{booktabs}
\usepackage{multirow}
\usepackage{geometry}
\usepackage{amsfonts,amssymb}

\usepackage{lineno,hyperref}
\modulolinenumbers[5]

\journal{arXiv}

\begin{document}

\begin{frontmatter}

\title{Multi-View representation learning in Multi-Task Scene}


\author[mymainaddress]{Run-kun Lu}

\author[mymainaddress]{Jian-wei Liu\corref{mycorrespondingauthor}}
\cortext[mycorrespondingauthor]{Corresponding author}
\ead{liujw@cup.edu.cn}

\author[mymainaddress]{Si-ming Lian}

\author[mymainaddress]{Xin Zuo}

\address[mymainaddress]{Department of Automation, College of Information Science and Engineering,
China University of Petroleum , Beijing, Beijing, China}

\begin{abstract}
Over recent decades have witnessed considerable progress in whether multi-task learning or multi-view learning, but the situation that consider both learning scenes simultaneously has received not too much attention. How to utilize multiple views’ latent representation of each single task to improve each learning task’s performance is a challenge problem. Based on this, we proposed a novel semi-supervised algorithm, termed as Multi-Task Multi-View learning based on Common and Special Features (MTMVCSF). In general, multi-views are the different aspects of an object and every view includes the underlying common or special information of this object. As a consequence, we will mine multiple views’ jointly latent factor of each learning task which consists of each view’s special feature and the common feature of all views. By this way, the original multi-task multi-view data has degenerated into multi-task data, and exploring the correlations among multiple tasks enables to make an improvement on the performance of learning algorithm. Another obvious advantage of this approach is that we get latent representation of the set of unlabeled instances by the constraint of regression task with labeled instances. The performance of classification and semi-supervised clustering task in these latent representations perform obviously better than it in raw data. Furthermore, an anti-noise multi-task multi-view algorithm called AN-MTMVCSF is proposed, which has a strong adaptability to noise labels. The effectiveness of these algorithms is proved by a series of well-designed experiments on both real world and synthetic data.
\end{abstract}

\begin{keyword}
multi-view, multi-task, latent representation, special feature, common feature
\end{keyword}

\end{frontmatter}

\section{Introduction}

In real world applications, different tasks can be associated with each other in various ways, and a task can be expressed from a variety of perspectives. For instance, the classification task on Facebook’s web pages correlated to the classification task on Twitter’s web pages. At the same time, we naturally define images and documents modals exist in web pages as two different views. Another example is an image data set called Leaves, which includes leaves’ photos of 32 genera, and the 32 genera can be viewed as 32 tasks. And the data provides three preprocessed features which can be viewed as three views.
Under mild assumptions, either multi-task or multi-view learning is more effective than single task and view algorithms. In general, multi-task learning enables to make an improvement on each task’s learning performance by utilizing the tasks’ correlations \cite{1zhang2016multi}, and multi-view learning usually utilize the different representations of raw data. More specifically, multi-view learning expects to find the consistency relationships among views, e.g. CCA based algorithms \cite{2chaudhuri2009multi,3kursun2010canonical}, or find the complementarity of various views, such as co-training \cite{4blum1998combining}. However, both consistency and complementarity information are meaningful, and ignoring any kind of them is a waste. Therefore, \cite{5liu2014partially} proposed PSLF and utilizes both the properties of consistency and complementarity among multiple views, which makes full use of the features of raw data. As a matter of fact, consistency corresponds to the concept’s consensus or agreement closely \cite{6brefeld2005multi,7zhou2010semi}, which is the common information of different views. On the contrary, complementarity corresponds to the concept’s diversity or disagreement \cite{8wang2007analyzing}, which is always the special information of each view.

In brief, multiple views can be viewed as different aspects of an object and they have the same label. Obviously, since multiple views represent various aspects of the same object, they may have shared some characteristics, and every view in multiple views should has its own unique characteristics, which is different from other ones. Thus, common and special features exist among different views, and we can make use of them to find a single-view joint latent feature representation to replace these views. More generally, in multi-task multi-view scene, multiple views exist in each task, which means we can get a single-view latent representation in each task. Clearly, with the latent representation in each task, the original multi-task multi-view data has degenerated into multi-task data. Simultaneously, in each task, by regularizing the weight of a regression task, we find the relationship among different tasks that also helps the division of latent feature representation. Furthermore, in classification or semi-supervised clustering learning scenes, only some parts of instances are labeled, which are valuable prior knowledge when classifying test samples or clustering unlabeled samples. Therefore, we will supervise the process of learning latent representation of unlabeled samples according to the training process of labeled samples. The above discussion is the main motivation of this paper, and we can summarize it as: it’s worth looking for each task’s latent representation fused by multiple views that considering agreement and disagreement among views simultaneously, and by considering the relationships of multiple tasks, not only can we improve the performance of learning task but also it helps the division of latent feature representation. 

Based on the discussion above, we proposed a semi-supervised algorithm, termed as Multi-Task Multi-View learning based on Common and Special Features (MTMVCSF). Different from other multi-task multi-view algorithms \cite{1zhang2016multi,9he2011graphbased,10zhang2012inductive,11qian2012reconstruction,12jin2013shared,13jin2014multi}, in the aspect of model structure, MTMVCSF takes both common and special information of multi-view data in each task into consideration, and the relationships among multiple tasks are utilized to make an improvement on learning task’s performance.

In some real-world applications, noise labels will seriously affect the performance of the model. As we known, when collecting data, we may mislabel some instances because of human’s negligence that some instances in data set may have wrong labels and we define this kind of labels as noise labels. According to our experimental results, MTMVCSF suffered from noises labels problem. Therefore, it is imperative to find an anti-noise algorithm to mitigate the adverse effects of noise labels. As far as we know, Zhibin Hong et al. \cite{14hong2013tracking} and Xue Mei et al. \cite{15mei2015robust} have exploited rMTFL proposed by Pinghua Gong et al. \cite{16gong2012robust} to achieve robust multi-task multi-view tracking in videos. However, rMTFL considers the outlier tasks, instead of outlier labels. Based on MTMVCSF, we proposed an anti-noise multi-task multi-view algorithm, termed as Anti-Noise MTMVCSF (AN-MTMVCSF). By adding a noise regression coefficient in objective function, the ability of the model to fight against noise labels becomes stronger.

The main contributions of this paper are summarized as follows:

(1) Different from algorithms that only consider views’ consistency \cite{1zhang2016multi,9he2011graphbased,10zhang2012inductive,12jin2013shared,13jin2014multi} or complementarity \cite{11qian2012reconstruction} in each learning task, MTMVCSF utilizes both of the properties. By considering these two factors simultaneously, we can learn more complete information contained in data and get better representation for each task;

(2) We reconstruct each task’s input data from different views to construct a single-view latent feature of each task, and this process is supervised by labeled samples. As a result, each task’s multi-view representations can be replaced by the corresponding joint single view latent feature. Therefore, the original multi-task multi-view scene has degenerated into multi-task scene;

(3) We proposed an anti-noise version algorithm termed as AN-MTMVCSF. In real world application, the presence of noise labels will seriously affect the performance of the model. Under these circumstances, MTMVCSF’s performance is not expected, however, with the addition of noise weight, AN-MTMVCSF shows superior performance and is substantially anti-noise to label corruption. In subsection 6.4, we verify that the ability of anti-noise of AN-MTMVCSF is much better than MTMVCSF.

The rest of the essay is organized as follows. In section 2, we overview some related work on multi-task and multi-view learning. In section 3, we define the notations and derive the objective functions of MTMVCSF. In section 4, we solve MTMVCSF’s update formulas. In section 5, we proposed AN-MTMVCSF and derived the corresponding update formulas. In section 6, we designed a substantial number of experiments to demonstrate the performance of the algorithms. In last section, we made a conclusion and discussed some future works about our proposed methods. 

\section{Related work}

\textbf{Multi-task Learning:} Recently, this typical kind of learning pattern has attracted significant research interest, and it is widely used in various fields. \cite{17gonccalves2016multi} proposed a multi-task model based on Gaussian copula model which considers tasks relationship structure’s joint estimation and each individual task parameters. Considering the exiting of outlier tasks, a robust multi-task learning method is proposed by Pinghua Gong et al. \cite{16gong2012robust}. And recently, low-rank and sparse method is proposed to find the common and special features among tasks \cite{18kong2017probabilistic}. Li Xiao et al. applied manifold regularized based multi-task learning into IQ prediction and achieved a great improvement compared with some benchmarks \cite{19xiao2019manifold}. Rajeev Ranjan et al. provided a deep multi-task algorithm that can deal with multiple visual tasks, such as face detection, gender recognition and etc. \cite{20ranjan2017hyperface}. Yiwei He et al. combine GANs and Multi-task learning together to handle gait recognition problem and achieve competitive results \cite{21he2018multi}.

\textbf{Multi-view Learning:} In real world application, some objects can be expressed into various perspectives naturally, others that do not have explicit multi-view representations may have latent ones, and both types of representations are conducive to find the data’s fundamental properties. \cite{2chaudhuri2009multi,3kursun2010canonical,22zhang2018generalized,23cheng2018tensor} utilize the agreement of multiple views, and \cite{4blum1998combining,24yin2018multiview} utilize the disagreement of multiple views. However, both consistency and complementarity play a significant role in multi-view learning. Therefore, \cite{5liu2014partially} proposed partially shared latent factor (PSLF) learning exploring the both properties of multi-view data, and on this basis, Zhong Zhang et al. \cite{25zhang2018multi} proposed a discriminative framework to modify PSLF. Chao and Sun proposed multi-view maximum entropy discrimination (MVMED) \cite{26sun2013multi}, which considers the concept of multi-view learning in the field of maximum entropy discrimination learning. And they further improved it  to an alternative version, termed as AMVMED by enforcing the posteriors of two view margins to be the same \cite{27chao2015alternative}. And multi-view learning is widely used in some daily applications, such as nature language process \cite{28fang2014detecting,29cui2018mv} and computer vision \cite{30cui2018mv,31ma2018learning,32huang2018deep}.

\textbf{Multi-task and Multi-view Learning:} As the name suggests, this is a kind of learning scene considering multi-task and multi-view simultaneously. We find that some multi-task multi-view learning frameworks proposed in \cite{1zhang2016multi,9he2011graphbased,10zhang2012inductive,11qian2012reconstruction,12jin2013shared,13jin2014multi,33bengio2009learning} only focus on the consistency of multiple views in each task. However, some new research works begin to realize that both consistency and complementarity are significant in such learning problem. For example, \cite{34lu2018multi} proposed a joint matrix factorization algorithm, and \cite{35wu2018dmtmv} proposed a deep learning based model to find shared and unified features of multiple views in each task. Additionally, \cite{36sun2018robust} proposed an online model using lifelong learning to find views’ representations of the coming task. Qian Zhang et al. provide a feature embedding method, which addresses the issue of learning an better feature embedding for the subsequent learning tasks \cite{37zhang2015mmfe}. Besides, multi-task and multi-view learning is widely used in some daily applications. For example, it is used in head-pose classiﬁcation \cite{38yan2013no,39yan2014evaluating}, infer user’s affective state \cite{40kandemir2014multi}, and video tracking problems \cite{14hong2013tracking,15mei2015robust}. And some recent works, such as \cite{41javanmardi2018robust,42han2017multi,43li2017fast} are also applied to the visual field.

\section{Framework}

\subsection{Notations}

In this paper, matrices are denoted by bold uppercase characters, vectors are denoted by bold lowercase characters, and other not bold characters stand for scalars. Additionally, Commas in square brackets means that the matrix or vector is connected by column, such as ${\bf{A}} = \left[ {{\bf{A}}_1 ,{\bf{A}}_2 } \right] \in ^{a \times b}$, where ${\bf{A}}_1  \in ^{a \times b_1 }$, ${\bf{A}}_2  \in ^{a \times b_2 }$, and $b=b_1+b_2$. On the contrary, semicolon in square brackets means that the matrix or vector is connected by row, such as ${\mathbf{B}} = \left[ {{\mathbf{B}}_1 ;{\mathbf{B}}_2 } \right] \in \mathbb{R}^{a \times b}$, where ${\mathbf{B}}_1  \in \mathbb{R}^{a_1  \times b}$, ${\mathbf{B}}_2  \in \mathbb{R}^{a_2  \times b}$, and $a=a_1+a_2$.

Supposed that ${\mathbf{X}}_t^v  \in \mathbb{R}^{M_t^v  \times N_t^v }$ is the set of input instances of view $v$ in task $t$, where $N_t^v$ is instances number, $M_t^v$ is the feature dimension. For convenience, we define different views and tasks’ instance numbers are the same, i.e.$N_t^v$ can be taken placed by N. Thus, in task t, the v-th instances set is:

\begin{equation}
	\label{eq1}
	{\mathbf{X}}_t^v  = \left[ {{\mathbf{x}}_{t,1}^v ,{\mathbf{x}}_{t,2}^v , \cdots ,{\mathbf{x}}_{t,i}^v , \cdots ,{\mathbf{x}}_{t,N}^v } \right] 	
\end{equation}
where ${\mathbf{x}}_{t,i}^v  \in \mathbb{R}^{M_t^v } ,i \in \{ 1, \cdots ,N\} $.

We divided data sets into labeled and unlabeled parts, and we denote the first $N_l$ instances of ${\mathbf{X}}_t^v$ are labeled, and the remaining $N_u$ instances are unlabeled, where $N_l+N_u=N$. On the other hand, assume that all the tasks have the same sample size N, and the same proportion of labeled or unlabeled samples. Thus, ${\mathbf{X}}_t^v $can be represented as ${\mathbf{X}}_t^v  = \left[ {{\mathbf{X}}_{t,l}^v ,{\mathbf{X}}_{t,u}^v } \right]$, where ${\mathbf{X}}_{t,l}^v  \in \mathbb{R}^{M_t^v  \times N_l }$, and ${\mathbf{X}}_{t,u}^v  \in \mathbb{R}^{M_t^v  \times N_u } $.

Following the above assumption, in each task, task label matrix   can be denoted as follows:

\begin{equation}
	\label{eq2}
{\mathbf{Y}}_t  = \left[ {{\mathbf{y}}_{t,1} ,{\mathbf{y}}_{t,2} , \cdots ,{\mathbf{y}}_{t,i} , \cdots ,{\mathbf{y}}_{t,N_l } } \right]
\end{equation}
where ${\mathbf{y}}_{t,i}  \in \mathbb{R}^C ,i \in \{ 1, \cdots ,N_l \} $,and there is a corresponding relationship between ${\mathbf{x}}_{t,i}^v $and ${\mathbf{y}}_{t,i} $.However, we need to remember that although multiple views exist in a task, they have the same label because different views in a task describe a same target. In TABLE I, we summarize the symbols and notations in this paper.

\begin{table}[!htbp]
	\centering
	\caption{Notation}
	\label{tb1}
	\begin{tabular}{ccc}
		\hline	
NOTATION	& SIZE & DESCRIPTION\\ \hline
$\mathbf{X}_t^v$  & $M_t^v\times N$ &the data matrix of $v$-th view in task $t$\\
${\mathbf{x}}_{t,i}^v$ & $M_t^v  \times 1$, $i \in \left\{ {1,\; \cdots ,N} \right\}$ &the $i$-th sample of the data of $v$-th view in task $t$\\
$\mathbf{X}_{t,l}^v$ & $M_t^v\times N_l$ &the labeled data matrix of $v$-th view in task $t$\\
$\mathbf{X}_{t,u}^v$ & $M_t^v\times N_u$ &the unlabeled data matrix of $v$-th view in task $t$\\
$\mathbf{Y}_t$ & $C\times N_l$ & the label matrix in task $t$\\
$\mathbf{y}_{t,i}$ & $C\times 1,i\in \{1,...,N\}$ &the $i$-th sample's label in task $t$\\
$\mathbf{B}_t^v$ & $M_t^v\times (K_s+K_c)$ &the basis matrix of $v$-th view in task $t$\\
$\mathbf{B}_t^{v,s}$ & $M_t^v\times K_s$ & the first $K_s$ columns of $\mathbf{B}_t^v$\\
$\mathbf{B}_t^{v,c}$ & $M_t^v\times K_c$ & the last $K_c$ columns of $\mathbf{B}_t^v$\\
$\mathbf{F}_t^v$ & $(K_s+K_c)\times N$ & the factor matrix of $v$-th view in task $t$\\
$\mathbf{F}_{t,l}^v$ & $(K_s+K_c)\times N_l$ & the labeled factor matrix of $v$-th view in task $t$\\
$\mathbf{F}_{t,u}^v$ & $(K_s+K_c)\times N_u$ & the unlabeled factor matrix of $v$-th view in task $t$\\
$\mathbf{F}_{t,l}^{v,s}$ & $K_s\times N_l$ & the first $K_s$ rows of $\mathbf{F}_{t,l}^v$\\
$\mathbf{F}_{t,u}^{v,s}$ & $K_s\times N_u$ & the first $K_s$ rows of $\mathbf{F}_{t,u}^v$\\
$\mathbf{F}_t^c$ & $K_c\times N$ & the last $K_c$ rows of $\mathbf{F}_t^v$, shared by $\mathbf{F}_t^1,...,\mathbf{F}_t^V$\\
$\mathbf{F}_{t,l}^c$ & $K_c\times N_l$ & the first $N_l$ columns of $\mathbf{F}_t^c$\\
$\mathbf{F}_{t,u}^c$ & $K_c\times N_u$ & the last $N_l$ columns of $\mathbf{F}_t^c$\\
$\mathbf{F}_t$ & $K\times N,K=K_s\times V+K_c$ & the matrix of required features in task $t$\\
$\mathbf{F}_{t,l}$ & $K\times N_l$ &the first $N_l$ columns of $\mathbf{F}_t^c$\\
$\mathbf{F}_{t,u}$ & $K\times N_u$ &the first $N_u$ columns of $\mathbf{F}_t^c$\\
$\mathbf{W}_t$ & $K\times C$ & the regression coefficient matrix in task $t$\\
$\mathbf{W}_d$ & $K\times C$ & the noise regression coefficient matrix\\
$\pi _t^v$ & $1\times 1$ & the weight of $v$-th view in task $t$\\
$\mathbf{\pi}$ & $1\times (T\times V)$ &the weight vector of different views in different task\\
$\mathbf{D}$ &$K\times K$ &a positive semi-definite matrix\\
 \hline
	\end{tabular}
\end{table}

In the following two subsections 3.2, 3.3, and section 4, we will introduce our proposed method from three major aspects respectively:

(1) Discussing the situation only consider the relationships among views in each task;

(2) Considering tasks’ relevance and deriving the joint objective function; 

(3) Solving the objective function and deriving the update formulas.

\subsection{Discussion about multi-view scene in a typical task}

$\mathbf{X}_t^v$ can be factored into the product of two nonnegative matrices: ${\mathbf{X}}_t^v  \leftarrow {\mathbf{B}}_t^v {\mathbf{F}}_t^v $,where ${\mathbf{B}}_t^v  = [{\mathbf{B}}_t^{v,s} ,{\mathbf{B}}_t^{v,c} ]$ is the matrix composed of basis vectors which project $\mathbf{X}_t^v$ into two subspaces, one is consistent space and the other is complementary one.$\mathbf{F}_t^v$ is the feature matrix which is composed of four parts: ${\mathbf{F}}_t^v  = \left[ {{\mathbf{F}}_{t,l}^{v,s} ,{\mathbf{F}}_{t,u}^{v,s} ;{\mathbf{F}}_{t,l}^c ,{\mathbf{F}}_{t,u}^c } \right]$. Among them, $\mathbf{F}_{t,l}^{v,s}$ is labeled and complementary factor; $\mathbf{F}_{t,u}^{v,s}$ is unlabeled and complementary factor. Note that we enforce consistent factor of different views to be equal, i.e. the consistent factor of different views is ${\mathbf{F}}_t^c  = \left[ {{\mathbf{F}}_{t,l}^c ,{\mathbf{F}}_{t,u}^c } \right]$, where ${\mathbf{F}}_{t,l}^c$ is the labeled and consistent factor, and ${\mathbf{F}}_{t,u}^c$ is the unlabeled and consistent factor.

After division, we will fuse the complementary features of different views and consistent feature in this task together using the rule: ${\mathbf{F}}_t  = \left[ {{\mathbf{F}}_t^{1,s} ;{\mathbf{F}}_t^{2,s} ; \cdots ;{\mathbf{F}}_t^{V,s} ;{\mathbf{F}}_t^c } \right]$.By this way, we construct a new matrix ${\mathbf{F}}_t  \in \mathbb{R}^{K \times N}$, which is actually a single view joint latent feature representation of task $t$, where $K = K_s  \times V + K_c $.We can substitute $\mathbf{F}_t$ for multiple views’ input instances sets of task $t$, which means $\mathbf{F}_t$ is viewed as the new instances set of task $t$ because it is each task’s high-level feature obtained through the fusion of multiple views. To illustrate this further, the above referred division and fusion process diagram is shown in Fig. 1.

\begin{figure}[!htbp]
	\centering
	\includegraphics[scale=1]{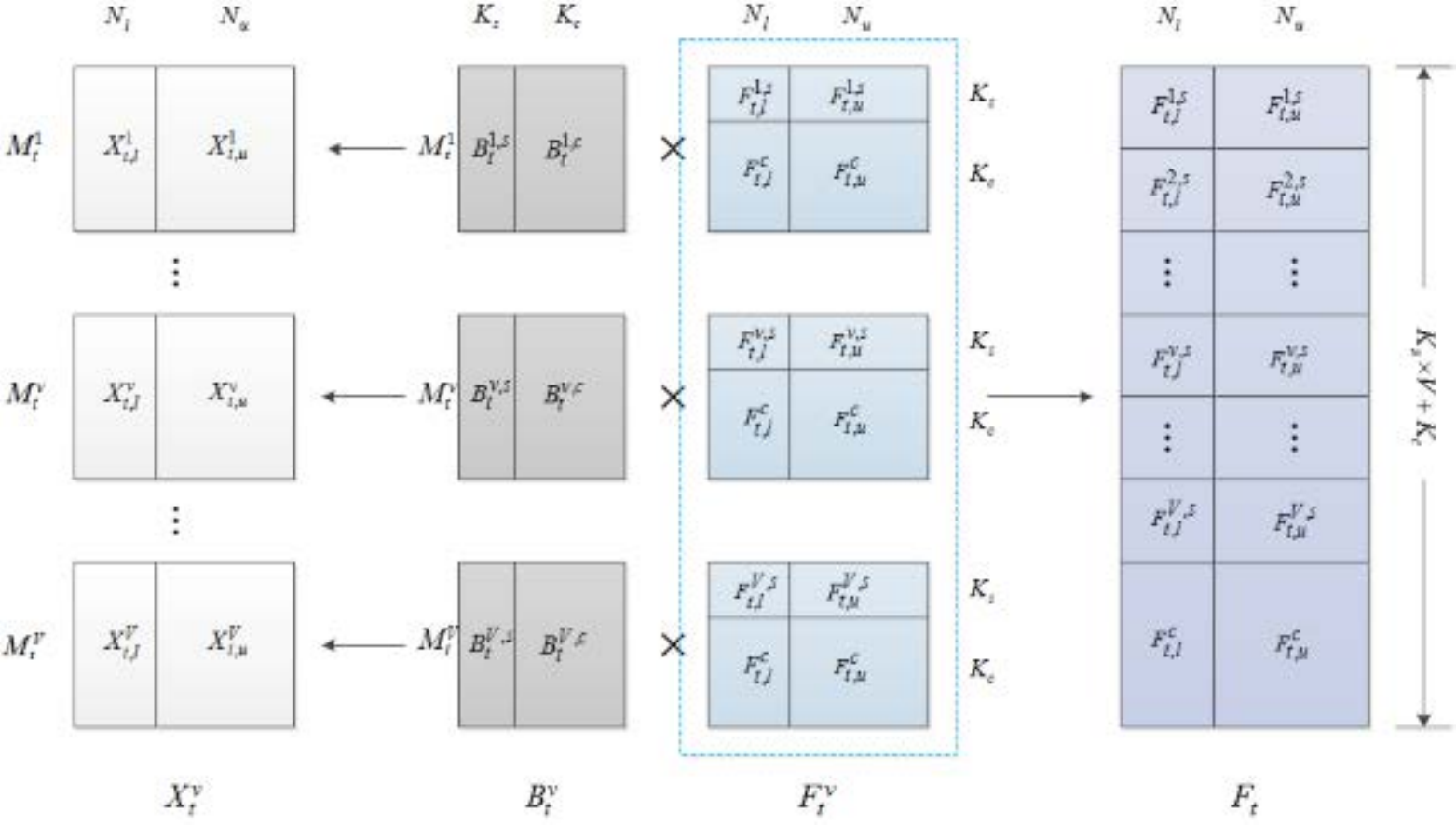}
	\caption{Division diagram of MTMVCSF in each task.}
	\label{fig1}
\end{figure}

We formulate the learning problem discussed above as that of minimizing the reconstruction error of the input instances coming from $V$ views and $T$ tasks. However, instances coming from different views may play a different role during the learning process. Therefore, we assign a weight $\pi_t^v$ to each view in each task, and intuitively, this weight is the measure of the importance of view. Let $\mathbf{\pi}$ the weight vector which is composed with all the weight vectors, i.e. ${\mathbf{\pi }} = (\pi _1^1 , \cdots ,\pi _1^V , \cdots ,\pi _t^1 , \cdots ,\pi _t^V , \cdots ,\pi _T^1 , \cdots ,\pi _T^V )$. As a consequence, the optimization problem is defined as follows:

\begin{equation}
	\label{eq3}
\begin{array}{l}
 \mathop {\min }\limits_{{\bf{B}}_t^v ,{\bf{F}}_t^v ,\pi _t^v } \sum\nolimits_{t = 1}^T {\sum\nolimits_{v = 1}^V {\pi _t^v \left\| {{\bf{X}}_t^v  - {\bf{B}}_t^v {\bf{F}}_t^v } \right\|_F^2  + \lambda \left\| {\bf{\pi }} \right\|} } _2^2  \\ 
 s.t.\;\;{\bf{B}}_t^v  \ge 0,{\bf{F}}_t^v  \ge 0,{\bf{\pi }} \ge 0,\sum\nolimits_{t = 1}^T {\sum\nolimits_{v = 1}^V {\pi _t^v }  = 1}  \\ 
 \end{array}
\end{equation}

\subsection{Scenario of different tasks}

${\bf{F}}_t  \in ^{K \times N} $is the new input instance set of task $t$, which is a single-view data, and it can be denoted as follow:

\begin{equation}
	\label{eq4}
{\bf{F}}_t  = ({\bf{f}}_{t,1} ,{\bf{f}}_{t,2} , \cdots ,{\bf{f}}_{t,i} , \cdots {\bf{f}}_{t,N} ),{\bf{f}}_{t,i}  \in ^{K \times 1} ,i \in [1, \cdots ,N]
\end{equation}
where the first $N_l$ columns of $\mathbf{F}_t$ are labeled instances and the remaining columns of $\mathbf{F}_t$ are unlabeled ones, i.e.${\bf{F}}_t  = ({\bf{F}}_{t,l} ,{\bf{F}}_{t,u})$, where  ${\bf{F}}_{t,l}  \in ^{K \times N_l } $,and ${\bf{F}}_{t,u}  \in ^{K \times N_u } $.

Assume that $\bf{Y}_t$ contains $N_l$ labels, because $\bf{F}_t$ is the joint latent representation of the set of input instances of different views in task $t$, and we use $\bf{F}_t$ in place of ${\bf{X}}_t^1 ,{\bf{X}}_t^2 , \cdots ,{\bf{X}}_t^V $as input instances set. As a consequence, we may define the estimating value as ${\bf{\hat Y}}_t  = {\bf{W}}_t^{\rm T} {\bf{F}}_{t,l}$, where ${\bf{W}}_t  \in ^{K \times C} $ is the weight matrix of task $t$ . Meanwhile, we concatenate every task’s weight matrices and obtain matrix ${\bf{W}} = \left[ {{\bf{W}}_1 ,{\bf{W}}_2 , \cdots ,{\bf{W}}_T } \right] \in ^{K \times \left( {C \times T} \right)} $.Moreover, we define ${\bf{W}} = {\bf{UA}}$, where ${\bf{U}} \in ^{K \times K} $ is a basis matrix that projects different tasks into a common subspace; and ${\bf{A}} = \left( {{\bf{A}}_1 ,{\bf{A}}_2 , \cdots ,{\bf{A}}_T } \right)$ is the set of matrix ${\bf{A}}_t  \in ^{K \times C} $,where ${\bf{A}}_t $ is task $t$’s regression coefficient matrix. Therefore,

\begin{equation}
	\label{eq5}
{\bf{\hat Y}}_t  = {\bf{W}}_t^{\rm T} {\bf{F}}_{t,l}  = ({\bf{UA}}_t )^{\rm T} {\bf{F}}_{t,l} 
\end{equation}

Drawing on the above analysis, we consider the following general regression problem \cite{44evgeniou2007multi}:

\begin{equation}
	\label{eq6}
\mathop {\min }\limits_{{\bf{A}}_t ,{\bf{U}}^{\rm T} } \sum\nolimits_{t = 1}^T {L\left( {{\bf{Y}}_t ,({\bf{UA}}_t )^{\rm T} {\bf{F}}_{t,l} } \right) + \gamma \left\| {\bf{A}} \right\|} _{2,1}^2 
\end{equation}
where $\left\|  \cdot  \right\|_{2,1} $ is the $L_{2,1}$-norm, for any ${\bf{M}} \in ^{I \times J}$, it is defined as $\left\| {\bf{M}} \right\|_{2,1}  = \sum\nolimits_{i = 1}^I {\sqrt {\sum\nolimits_{j = 1}^J {{\bf{M}}_{ij}^2 } } } $.And $L_{2,1}$-norm ensures the sparse solution of regression coefficient matrix and combines the features of tasks to ensure that the related features will be selected across them. The above discussion is based on the assumption that there are only partial shared features exist among different tasks, that is to say, $\bf{A}$ has many zero rows, and therefore the corresponding columns of matrix $\bf{U}$ will not work.

Equation (6) is a non-convex optimization learning problem, and   term is non-smooth. Both of the reasons cause the problem to be inconvenient to optimize, and therefore \cite{44evgeniou2007multi} transformed the original problem into an equivalent form:   

\begin{equation}
	\label{eq7}
\mathop {\min }\limits_{{\bf{W}}_t ,{\bf{D}}} \sum\nolimits_{t = 1}^T {L({\bf{Y}}_t ,{\bf{W}}_t^{\rm T} {\bf{F}}_{t,l} ) + \gamma \sum\nolimits_{t = 1}^T {trace({\bf{W}}_t^{\rm T} {\bf{D}}^ +  {\bf{W}}_t )} } 
\end{equation}
where ${\bf{D}} \in ^{K \times K} $ is a positive semi-definite matrix, and we definite it as ${\bf{D}} = \left( {{\bf{WW}}^{\rm T} } \right)^{1/2} /trace\left( {{\bf{WW}}^{\rm T} } \right)^{1/2} $ according to \cite{44evgeniou2007multi}, ${\bf{D}}^ +  $ is its pseudo-inverse, and $gamma$ is a tradeoff hyperparameter.

We have now reached the position to explicit give a joint optimization problem considering the situation contains both multi-task and multi-view according to equations (3) and (7):

\begin{equation}
	\label{eq8}
\begin{array}{l}
 \mathop {\min }\limits_{{\bf{B}}_t^v ,{\bf{F}}_t^v ,\pi _t^v ,{\bf{W}}_t ,{\bf{D}}} \sum\nolimits_{t = 1}^T {\sum\nolimits_{v = 1}^V {\pi _t^v \left\| {{\bf{X}}_t^v  - {\bf{B}}_t^v {\bf{F}}_t^v } \right\|_F^2  + \lambda \left\| {\bf{\pi }} \right\|} } _2^2  + \beta (\sum\nolimits_{t = 1}^T {L({\bf{Y}}_t ,{\bf{W}}_t^{\rm T} {\bf{F}}_{t,l} ) + \gamma \sum\nolimits_{t = 1}^T {trace({\bf{W}}_t^{\rm T} {\bf{D}}^ +  {\bf{W}}_t )} } ) \\ 
 s.t.\;\;{\bf{B}}_t^v  \ge 0,{\bf{F}}_t^v  \ge 0,{\bf{\pi }} \ge 0,\sum\nolimits_{t = 1}^T {\sum\nolimits_{v = 1}^V {\pi _t^v }  = 1}  \\ 
 \end{array} 
\end{equation}
where $\beta$ is a nonnegative tradeoff hyperparameter.

Equation (8) is an overall objective function which contains several sub-objective functions, and I will make a summarize of them: the first term in Equation (8) aims to minimize the reconstruction error of non-negative data, and obviously ${\bf{B}}_t^v  \ge 0,{\bf{F}}_t^v  \ge 0$ are constraints of it; the second term in Equation (8) is the weights’ regularization of first term, and ${\bf{\pi }} \ge 0,\sum\nolimits_{t = 1}^T {\sum\nolimits_{v = 1}^V {\pi _t^v }  = 1}$ are constraints of it; the third term in Equation (8) is the prediction error between true labels and predictive labels; and the last term in Equation (8) is the regularization term that combines different tasks together and utilizes the consistency of each task.

\section{Derivation of Updating Formula}

In this section, we develop the learning algorithm for joint optimization problem (8). Note that there are many variants for loss function in problem (8), such as least square, KL divergence, Bregman distance, and so on. Without loss of generality and to make our presentation easy to understand, here we will give the results of linear regression:

\begin{equation}
	\label{eq9}
Z = \sum\nolimits_{t = 1}^T {\sum\nolimits_{v = 1}^V {\pi _t^v \left\| {{\bf{X}}_t^v  - {\bf{B}}_t^v {\bf{F}}_t^v } \right\|_F^2  + \lambda \left\| {\bf{\pi }} \right\|} } _2^2  + \beta (\sum\nolimits_{t = 1}^T {\left\| {{\bf{Y}}_t  - {\bf{W}}_t^{\rm T} {\bf{F}}_{t,l} } \right\|_F^2  + \gamma \sum\nolimits_{t = 1}^T {trace({\bf{W}}_t^{\rm T} {\bf{D}}^ +  {\bf{W}}_t )} } ) 
\end{equation}

\subsection{Updating rule for $\bf{W}_t$}

Fixing $\bf{F}_t$ and $\bf{D}$, the derivative of objective function about $\bf{W}_t$ is:

\begin{equation}
	\label{eq10}
\frac{{\partial Z}}{{\partial {\bf{W}}_t }} = 2\beta [{\bf{F}}_{t,l} ({\bf{W}}_t^{\rm T} {\bf{F}}_{t,l}  - {\bf{Y}}_t )^{\rm T}  + \gamma ({\bf{D}}^{ - 1}  + ({\bf{D}}^{ - 1} )^{\rm T} ){\bf{W}}_t ]
\end{equation}

Let ${\bf{DD}} = {\bf{D}}^{ - 1}  + ({\bf{D}}^{ - 1} )^{\rm T} $ , and setting $\partial Z/\partial {\bf{W}}_t  = 0$ , we can get the following updating rule:

\begin{equation}
	\label{eq11}
{\bf{W}}_t  = ({\bf{F}}_{t,l} {\bf{F}}_{t,l}^{\rm T}  + \gamma {\bf{DD}})^{ - 1} {\bf{F}}_{t,l} {\bf{Y}}_t^{\rm T} 
\end{equation}

\subsection{Updating rule for $\bf{D}$}

Andreas Argyriou et al. have proposed an algorithm to update the value of $\bf{D}$ \cite{44evgeniou2007multi}:

\begin{equation}
	\label{eq12}
{\bf{D}} = ({\bf{WW}}^{\rm T} )^{1/2} /trace({\bf{WW}}^{\rm T} )^{1/2} 
\end{equation}
and the optimal value equals to $(trace({\bf{WW}}^{\rm T} )^{1/2} )^2 $. However, equation (11) will present some possible singularities on computing the inverse if we use equation (12), and we can solve this problem by using equation (12)’s optimal value $(trace({\bf{WW}}^{\rm T} )^{1/2} )^2 $\cite{44evgeniou2007multi}.

\subsection{Updating rule for ${\bf{B}}_t^v$}

Considering (9), the optimization function matrices ${\bf{B}}_t^v$ has the constraint condition ${\bf{B}}_t^v  \ge 0$. thus Lagrange multiplier method is used to solve the constraint problem. Lagrange dual function can be defined as follows:

\begin{equation}
	\label{eq13}
L({\bf{B}}_t^v ) = \sum\nolimits_{t = 1}^T {\sum\nolimits_{v = 1}^V {\pi _t^v \left\| {{\bf{X}}_t^v  - {\bf{B}}_t^v {\bf{F}}_t^v } \right\|_F^2  - trace\left( {\Psi _b^{\rm T} {\bf{B}}_t^v } \right)} } 
\end{equation}
where $\Psi _b$ is the Lagrange multiplier, and the derivative of $L({\bf{B}}_t^v )$ with respect to ${\bf{B}}_t^v$ is:

\begin{equation}
	\label{eq14}
\frac{{\partial L({\bf{B}}_t^v )}}{{\partial {\bf{B}}_t^v }} = 2\pi _t^v ( - {\bf{X}}_t^v ({\bf{F}}_t^v )^{\rm T}  + {\bf{B}}_t^v {\bf{F}}_t^v ({\bf{F}}_t^v )^{\rm T} ) - \Psi _b 
\end{equation}

Let (14) equals to zero, and do dot multiply both sides of the equation by ${\bf{B}}_t^v $. According to Karush-Kuhn-Tucker (KKT) condition, and we get the following identity:

\begin{equation}
	\label{eq15}
({\bf{B}}_t^v {\bf{F}}_t^v ({\bf{F}}_t^v )^{\rm T} )_{mk} ({\bf{B}}_t^v )_{mk}  = ({\bf{X}}_t^v ({\bf{F}}_t^v )^{\rm T} )_{mk} ({\bf{B}}_t^v )_{mk} 
\end{equation}

And the updating rule w.r.t $\left( {{\bf{B}}_t^v } \right)_{mk}$ is:

\begin{equation}
	\label{eq16}
\left( {{\bf{B}}_t^v } \right)_{mk}  \leftarrow \left( {{\bf{B}}_t^v } \right)_{mk} \frac{{\left( {{\bf{X}}_t^v \left( {{\bf{F}}_t^v } \right)^{\rm T} } \right)_{mk} }}{{\left( {{\bf{B}}_t^v {\bf{F}}_t^v \left( {{\bf{F}}_t^v } \right)^{\rm T} } \right)_{mk} }} 
\end{equation}

\subsection{Updating rule for ${\bf{F}}_t^v$}

According to the definition in subsection 3.2, ${\bf{F}}_t^v$ is composed of four parts, and therefore we should update the formula from four aspects: i.e. computing the derivative of ${\bf{F}}_{t,l}^{v,s} $,${\bf{F}}_{t,l}^c $,${\bf{F}}_{t,u}^{v,s} $ and ${\bf{F}}_{t,u}^c $. We define $\Psi _t $ is the Lagrange multiplier corresponding to constraint ${\bf{F}}_t^v  \ge 0$, where $\Psi _t  = [\Psi _{t,l}^{v,s} ,\Psi _{t,u}^{v,s} ;\Psi _{t,l}^c ,\Psi _{t,u}^c ] = [(\Psi _t )_{kn} ]$. Using Lagrange multiplier method, we get the Lagrange function as follows:

\begin{equation}
	\label{eq17}
L({\bf{F}}_t^v ) = \sum\nolimits_{t = 1}^T {\sum\nolimits_{v = 1}^V {\pi _t^v \left\| {{\bf{X}}_t^v  - {\bf{B}}_t^v {\bf{F}}_t^v } \right\|_F^2 }  + } \beta \sum\nolimits_{t = 1}^T {\left\| {{\bf{Y}}_t  - {\bf{W}}_t^{\rm T} {\bf{F}}_{t,l} } \right\|_F^2 }  - trace(\Psi _t^{\rm T} {\bf{F}}_t^v )
\end{equation}

Compute the derivatives of the Lagrange function (17) with respect to ${\bf{F}}_{t,l}^{v,s} $,${\bf{F}}_{t,l}^c $,${\bf{F}}_{t,u}^{v,s} $,${\bf{F}}_{t,u}^c $ and let each of them equal to zero:

\begin{equation}
	\label{eq18}
\left\{ \begin{array}{l}
 \frac{{\partial L\left( {{\bf{F}}_{t,l}^{v,s} } \right)}}{{\partial {\bf{F}}_{t,l}^{v,s} }} = 2\pi _t^v \left( {\left( { - {\bf{B}}_{t,s}^v } \right)^{\rm T} \left( {{\bf{X}}_{t,l}^v  - {\bf{B}}_t^v {\bf{F}}_{t,l}^v } \right)} \right) + 2\beta H_t^{v,s}  - \Psi _{t,l}^{v,s}  = 0 \\ 
 \frac{{\partial L\left( {{\bf{F}}_{t,l}^c } \right)}}{{\partial {\bf{F}}_{t,l}^c }} = \sum\nolimits_{v = 1}^V {2\pi _t^v \left( {\left( { - {\bf{B}}_{t,c}^v } \right)^{\rm T} \left( {{\bf{X}}_{t,l}^v  - {\bf{B}}_t^v {\bf{F}}_{t,l}^v } \right)} \right) + 2\beta H_t^c  - \Psi _{t,l}^c  = 0}  \\ 
 \frac{{\partial L\left( {{\bf{F}}_{t,u}^{v,s} } \right)}}{{\partial {\bf{F}}_{t,u}^{v,s} }} = 2\pi _t^v \left( {\left( { - {\bf{B}}_{t,s}^v } \right)^{\rm T} \left( {{\bf{X}}_{t,u}^v  - {\bf{B}}_t^v {\bf{F}}_{t,u}^v } \right)} \right) - \Psi _{t,u}^{v,s}  = 0 \\ 
 \frac{{\partial L\left( {{\bf{F}}_{t,u}^c } \right)}}{{\partial {\bf{F}}_{t,u}^c }} = \sum\nolimits_{v = 1}^V {2\pi _t^v \left( {\left( { - {\bf{B}}_{t,c}^v } \right)^{\rm T} \left( {{\bf{X}}_{t,u}^v  - {\bf{B}}_t^v {\bf{F}}_{t,u}^v } \right)} \right) - \Psi _{t,u}^c  = 0}  \\ 
 \end{array} \right.
\end{equation}
where ${\bf{H}}_t  = {\bf{W}}_t {\bf{W}}_t^{\rm T} {\bf{F}}_{t,l}  - {\bf{W}}_t {\bf{Y}}_t  = [H_t^{1,s} ;H_t^{2,s} ; \cdots ;H_t^{V,s} ;H_t^c ]$.

Next, do the dot multiply both sides of the equations in (18) by ${\bf{F}}_{t,l}^{v,s} $,${\bf{F}}_{t,l}^c $,${\bf{F}}_{t,u}^{v,s} $,${\bf{F}}_{t,u}^c $, where $(\Psi _{t,l}^{v,s} )_{kn} ({\bf{F}}_{t,l}^{v,s} )_{kn}  = 0$,$(\Psi _{t,l}^c )_{kn} ({\bf{F}}_{t,l}^c )_{kn}  = 0$,$(\Psi _{t,u}^{v,s} )_{kn} ({\bf{F}}_{t,u}^{v,s} )_{kn}  = 0$,$(\Psi _{t,u}^c )_{kn} ({\bf{F}}_{t,u}^c )_{kn}  = 0$  according to KKT condition, and we get the following equation:

\begin{equation}
	\label{eq19}
\left\{ \begin{array}{l}
 \left( {{\bf{F}}_{t,l}^{v,s} } \right)_{kn}  \leftarrow \left( {{\bf{F}}_{t,l}^{v,s} } \right)_{kn} \left( {{\bf{Sq}}_{t,l}^{v,s} } \right)_{kn}  \\ 
 \left( {{\bf{F}}_{t,l}^c } \right)_{kn}  \leftarrow \left( {{\bf{F}}_{t,l}^c } \right)_{kn} \left( {{\bf{Sq}}_{t,l}^c } \right)_{kn}  \\ 
 \left( {{\bf{F}}_{t,u}^{v,s} } \right)_{kn}  \leftarrow \left( {{\bf{F}}_{t,u}^{v,s} } \right)_{kn} \left( {{\bf{Sq}}_{t,u}^{v,s} } \right)_{kn}  \\ 
 \left( {{\bf{F}}_{t,u}^c } \right)_{kn}  \leftarrow \left( {{\bf{F}}_{t,u}^c } \right)_{kn} \left( {{\bf{Sq}}_{t,u}^c } \right)_{kn}  \\ 
 \end{array} \right.
\end{equation}
where $\left( {{\bf{Sq}}_{t,l}^{v,s} } \right)_{kn}$,$\left( {{\bf{Sq}}_{t,l}^c } \right)_{kn}$,$\left( {{\bf{Sq}}_{t,u}^{v,s} } \right)_{kn}$,$\left( {{\bf{Sq}}_{t,u}^c } \right)_{kn}$ are defined as follows:

\begin{equation}
	\label{eq20}
\begin{array}{l}
 \left( {{\bf{Sq}}_{t,l}^{v,s} } \right)_{kn}  = \sqrt {\frac{{\left[ {\pi _t^v \left( {{\bf{B}}_{t,s}^v } \right)^{\rm T} {\bf{X}}_{t,l}^v  + \beta \left( {H_ +  } \right)_t^{v,s} } \right]_{kn} }}{{\left[ {\pi _t^v \left( {{\bf{B}}_{t,s}^v } \right)^{\rm T} {\bf{B}}_t^v {\bf{F}}_{t,l}^v  + \beta \left( {H_ -  } \right)_t^{v,s} } \right]_{kn} }}}  \\ 
 \left( {{\bf{Sq}}_{t,l}^c } \right)_{kn}  = \sqrt {\frac{{\left[ {\sum\nolimits_{v = 1}^V {\pi _t^v \left( {{\bf{B}}_{t,c}^v } \right)^{\rm T} {\bf{X}}_{t,l}^v  + \beta \left( {H_ +  } \right)_t^c } } \right]_{kn} }}{{\left[ {\sum\nolimits_{v = 1}^V {\pi _t^v \left( {{\bf{B}}_{t,c}^v } \right)^{\rm T} {\bf{B}}_t^v {\bf{F}}_{t,l}^v  + \beta \left( {H_ -  } \right)_t^c } } \right]_{kn} }}}  \\ 
 \left( {{\bf{Sq}}_{t,u}^{v,s} } \right)_{kn}  = \sqrt {\frac{{\left[ {\pi _t^v \left( {{\bf{B}}_{t,s}^v } \right)^{\rm T} {\bf{X}}_{t,u}^v } \right]_{kn} }}{{\left[ {\pi _t^v \left( {{\bf{B}}_{t,s}^v } \right)^{\rm T} {\bf{B}}_t^v {\bf{F}}_{t,u}^v } \right]_{kn} }}}  \\ 
 \left( {{\bf{Sq}}_{t,u}^c } \right)_{kn}  = \sqrt {\frac{{\left[ {\sum\nolimits_{v = 1}^V {\pi _t^v \left( {{\bf{B}}_{t,c}^v } \right)^{\rm T} {\bf{X}}_{t,u}^v } } \right]_{kn} }}{{\left[ {\sum\nolimits_{v = 1}^V {\pi _t^v \left( {{\bf{B}}_{t,c}^v } \right)^{\rm T} {\bf{B}}_t^v {\bf{F}}_{t,u}^v } } \right]_{kn} }}}  \\ 
 \end{array}
\end{equation}

where the matrices $\left( {{\bf{H}}_ +  } \right)_t $ and $\left( {{\bf{H}}_ -  } \right)_t $ are defined as follows:

\begin{equation}
	\label{eq21}
\begin{array}{l}
 \left( {{\bf{H}}_ +  } \right)_t  = \left( {{\bf{W}}_t {\bf{W}}_t^{\rm T} } \right)^ +  {\bf{F}}_{t,l}  + \left( {{\bf{W}}_t } \right)^ -  {\bf{Y}}_t  \\ 
 \;\;\;\;\;\;\;\; = [\left( {H_ +  } \right)_t^{1,s} ;\left( {H_ +  } \right)_t^{2,s} ; \cdots ;\left( {H_ +  } \right)_t^{V,s} ;\left( {H_ +  } \right)_t^c ] \\ 
 \left( {{\bf{H}}_ -  } \right)_t  = \left( {{\bf{W}}_t {\bf{W}}_t^{\rm T} } \right)^ -  {\bf{F}}_{t,l}  + \left( {{\bf{W}}_t } \right)^ +  {\bf{Y}}_t  \\ 
 \;\;\;\;\;\;\;\; = [\left( {H_ -  } \right)_t^{1,s} ;\left( {H_ -  } \right)_t^{2,s} ; \cdots ;\left( {H_ -  } \right)_t^{V,s} ;\left( {H_ -  } \right)_t^c ] \\ 
 \end{array}
\end{equation}

We define ${\bf{M}}^ +   = \left( {\left| {\bf{M}} \right| + {\bf{M}}} \right)/2$ and ${\bf{M}}^ -   = \left( {\left| {\bf{M}} \right| - {\bf{M}}} \right)/2$, where $\left| {\bf{M}} \right|$ is an operator computing the absolute value of each element in matrix $\bf{M}$.

\subsection{Updating rule for $\bf\pi$}

Fixing ${\bf{B}}_t^v $ and ${\bf{F}}_t^v $, the minimization problem of $Z$ is degenerated to a simple optimial problem w.r.t $\bf\pi$:

\begin{equation}
	\label{eq22}
\begin{array}{l}
 \mathop {\min }\limits_{\pi _t^v } \sum\nolimits_{t = 1}^T {\sum\nolimits_{v = 1}^V {f_t^v \pi _t^v  + \lambda \left\| {\bf{\pi }} \right\|_2^2 } }  \\ 
 s.t.\;{\bf{\pi }} \ge 0,\;\sum\nolimits_{t = 1}^T {\sum\nolimits_{v = 1}^V {\pi _t^v  = 1} }  \\ 
 \end{array}
\end{equation}
where $f_t^v  = \left\| {{\bf{X}}_t^v  - {\bf{B}}_t^v {\bf{F}}_t^v } \right\|_F^2 $ is a constant. We can solve the convex optimization with CVX\footnote{http://cvxr.com/cvx/} or fmincon\footnote{http://www.mathworks.com/help/optim/ug/fmincon.html} , where CVX is a MATLAB toolkit for convex optimization, and fmincon is a MATLAB function to solve to the minimum value of multiple linear or nonlinear function. 

The above analysis leads to the following Algorithm 1.

\begin{algorithm}
	\renewcommand{\algorithmicrequire}{\textbf{Input:}}
	\renewcommand{\algorithmicensure}{\textbf{Output:}}
	\caption{}
	\label{alg1}
	\begin{algorithmic}[1]
		\REQUIRE  the instances sets of $V$ views in $T$ tasks: ${\bf{X}}_1^1 ,{\bf{X}}_1^2 , \cdots ,{\bf{X}}_1^V , \cdots ,{\bf{X}}_t^1 ,{\bf{X}}_t^2 , \cdots {\bf{X}}_t^V , \cdots ,{\bf{X}}_T^1 ,{\bf{X}}_T^2 , \cdots ,{\bf{X}}_T^V $; the label matrices of $t$ tasks: ${\bf{Y}}_1 ,{\bf{Y}}_2 , \cdots ,{\bf{Y}}_T$; the parameters: $\lambda$, $\beta$, $\gamma$.
		\ENSURE The feature representation matrices:${\bf{F}}_1 ,{\bf{F}}_2 , \cdots ,{\bf{F}}_T $; the weight matrices: ${\bf{W}}_1 ,{\bf{W}}_2 , \cdots ,{\bf{W}}_T $.\\
		\textbf{Initialization: } The basis matrix ${\bf{B}}_t^v $ of different views in each task; the feature matrix ${\bf{F}}_t^v $ of different views in each task; the weight matrices: ${\bf{W}}_1 ,{\bf{W}}_2 , \cdots ,{\bf{W}}_T $; the positive semi-definite matrix ${\bf{D}}$; $\pi _t^v $ in vector ${\bf{\pi }}$ of different views in each tasks equal to $1/VT$.
		
		\textbf{Loop:}\\
		\textbf{For} each epoch \textbf{do:}\\
		\STATE Initialize ${\bf{B}}_t^v $,${\bf{F}}_t^v $,${\bf{W}}_1 ,{\bf{W}}_2 , \cdots ,{\bf{W}}_T $,${\bf{D}}$,${\bf{\pi }}$;
		\STATE \textbf{For} each task $t \in \left\{ {1, \cdots ,T} \right\}$ \textbf{do}:\\
		\qquad Update ${\bf{W}}_t $ using (11);
		\STATE Update ${\bf{D}}$ using (12);
		\STATE \textbf{For} each view $v \in \left\{ {1, \cdots ,V} \right\}$ \textbf{do}:\\
		\qquad \textbf{For} each task $t \in \left\{ {1, \cdots ,T} \right\}$ \textbf{do}:\\
		\qquad\qquad Update $\bf{B}_t^v$ using (16);\\
		\qquad\qquad Update ${\bf{F}}_{t,l}^{v,s} ,\;{\bf{F}}_{t,l}^c ,\;{\bf{F}}_{t,u}^{v,s} ,\;{\bf{F}}_{t,u}^c $ using (19);\\
		\qquad\qquad Update ${\bf{\pi }}$ using (22);
		\STATE ${\bf{F}}_t  \leftarrow \left[ {{\bf{F}}_{t,l}^{1,s} ,{\bf{F}}_{t,u}^{1,s} ; \cdots ;{\bf{F}}_{t,l}^{V,s} ,{\bf{F}}_{t,u}^{V,s} ;{\bf{F}}_{t,l}^c ,{\bf{F}}_{t,u}^c } \right]$
	\end{algorithmic}
\end{algorithm}

We should note that equation (11) ${\mathbf{W}}_t  = ({\mathbf{F}}_{t,l} {\mathbf{F}}_{t,l}^T  + \gamma {\mathbf{DD}})^{ - 1} {\mathbf{F}}_{t,l} {\mathbf{Y}}_t^T $ involves the matrix inverse calculation, which may be a disaster for large scale data. However, we find that $\bf{F}_{t,l}\bf{F}_{t,l}^T+\gamma \bf{DD}\in \mathbb{R}^{K\times K}$ ,and $K$ is hyper-parameter that the choice of it is shown in Table 3. In most cases, $K$ is 20, and sometimes it may be bigger but usually no more than 50. Because when $K$ is too small, the performance is bad because the subspace loose too much information; and when $K$ is too big, the performance may not get better and even get worse because the subspace may contain redundant information. Such computational complexity of the matrix inverse calculation is acceptable, and the computational complexity will not increase when meet high dimensional data.

\section{AN-MTMVCSF}

Label noise is an important challenge in semi-supervised learning, with many potential negative effects on the performance. By adding the noise weight in object function, we propose an anti-noise framework based on MTMVCSF called AN-MTMVCSF. According to formula (8), we add a noise weight term ${\bf{W}}_d$ as follows:

\begin{equation}
	\label{eq23}
\begin{array}{l}
 \mathop {\min }\limits_{{\bf{B}}_t^v ,{\bf{F}}_t^v ,{\bf{W}}_t ,{\bf{W}}_d ,{\bf{D}},\pi _t^v } \sum\nolimits_{t = 1}^T {\sum\nolimits_{v = 1}^V {\pi _t^v \left\| {{\bf{X}}_t^v  - {\bf{B}}_t^v {\bf{F}}_t^v } \right\|_F^2  + \lambda \left\| {\bf{\pi }} \right\|_2^2 } }  \\ 
  + \beta \left( {\sum\nolimits_{t = 1}^T {L\left( {{\bf{Y}}_t ,\left( {{\bf{W}}_t^{\rm T} {\rm{ + }}{\bf{W}}_d } \right){\bf{F}}_{t,l} } \right)}  + \mu \left\| {{\bf{W}}_d } \right\|_{2,1}^2  + \gamma \sum\nolimits_{t = 1}^T {trace({\bf{W}}_t^{\rm T} {\bf{D}}^ +  {\bf{W}}_t )} } \right) \\ 
 s.t.\;{\bf{B}}_t^v  \ge 0,\;{\bf{F}}_t^v  \ge 0,\;{\bf{\pi }} \ge 0,\;\sum\nolimits_{t = 1}^T {\sum\nolimits_{v = 1}^V {\pi _t^v  = 1} }  \\ 
 \end{array}
\end{equation}

In each task, the model parameter equals to the sum of ${\bf{W}}_t^{\rm T}$ and ${\bf{W}}_d$ , where the regularization term on ${\bf{W}}_t^{\rm T}$ captures the shared features among tasks, and the second term on ${\bf{W}}_d$ will add some disturbance for model, $L_{2,1}$-norm makes ${\bf{W}}_d$’s certain rows be zero and other rows be non-zero. The zero rows mean those tasks have no noise label problems and non-zero rows mean those tasks have noise label problems. Due to adding noise in model parameter, ${\bf{W}}_t^{\rm T}$ must have stronger capability to fight against noise. Simultaneously, the update progress of ${\bf{W}}_d$ also drive it to make smaller influence on learning process. Similar to the deduction in the previous section, the updating formulas of each parameter are shown as follows:

(1) Updating rule for ${\bf{W}}_t $:

\begin{equation}
	\label{eq24}
{\bf{W}}_t  = \left( {{\bf{F}}_{t,l} {\bf{F}}_{t,l}^{\rm T}  + \gamma {\bf{DD}}} \right)^{ - 1} \left( {{\bf{F}}_{t,l} {\bf{Y}}_t^{\rm T}  - {\bf{F}}_{t,l} {\bf{F}}_{t,l}^{\rm T} {\bf{W}}_d } \right)
\end{equation}

(2) Updating rule for ${\bf{W}}_d $:

\begin{equation}
	\label{eq25}
{\bf{W}}_d  = \left( {\sum\nolimits_{t = 1}^T {{\bf{F}}_{t,l} {\bf{F}}_{t,l}^{\rm T} }  + \mu {\bf{E}}_{\bf{d}} } \right)^{ - 1} \left( {\sum\nolimits_{t = 1}^T {{\bf{F}}_{t,l} {\bf{Y}}_t^{\rm T} }  - \sum\nolimits_{t = 1}^T {{\bf{F}}_{t,l} {\bf{F}}_{t,l}^{\rm T} {\bf{W}}_t } } \right)
\end{equation}
where ${\bf{E}}_d  \in ^{K \times K}$ is a diagonal matrix with its diagonal elements equal to $e\left( {K,K} \right) = \frac{1}{2}\left\| {{\bf{W}}_d \left( {k, \cdot } \right)} \right\|_2$.

(3) Updating rule for ${\bf{D}}$:

\begin{equation}
	\label{eq26}
{\bf{D}} = \left( {{\bf{WW}}^{\rm T} } \right)^{1/2} /trace\left( {{\bf{WW}}^{\rm T} } \right)^{1/2} 
\end{equation}
where ${\mathbf{W}} $ is consisting of ${\mathbf{W}}_t $ from all tasks, i.e.${\bf{W}} = [{\bf{W}}_1 ,{\bf{W}}_2 , \cdots ,{\bf{W}}_T ]$

(4) Updating rule of ${\bf{B}}_t^v$:

\begin{equation}
	\label{eq27}
\left( {{\bf{B}}_t^v } \right)_{mk}  \leftarrow \left( {{\bf{B}}_t^v } \right)_{mk} \frac{{\left( {{\bf{X}}_t^v \left( {{\bf{F}}_t^v } \right)^{\rm T} } \right)_{mk} }}{{\left( {{\bf{B}}_t^v {\bf{F}}_t^v \left( {{\bf{F}}_t^v } \right)^{\rm T} } \right)_{mk} }}
\end{equation}

(5) Updating rule of ${\bf{F}}_l^v$:

\begin{equation}
	\label{eq28}
\left\{ \begin{array}{l}
 \left( {{\bf{F}}_{t,l}^{v,s} } \right)_{kn}  \leftarrow \left( {{\bf{F}}_{t,l}^{v,s} } \right)_{kn} \left( {{\bf{Sq}}_{t,l}^{v,s} } \right)_{kn}  \\ 
 \left( {{\bf{F}}_{t,l}^c } \right)_{kn}  \leftarrow \left( {{\bf{F}}_{t,l}^c } \right)_{kn} \left( {{\bf{Sq}}_{t,l}^c } \right)_{kn}  \\ 
 \left( {{\bf{F}}_{t,u}^{v,s} } \right)_{kn}  \leftarrow \left( {{\bf{F}}_{t,u}^{v,s} } \right)_{kn} \left( {{\bf{Sq}}_{t,u}^{v,s} } \right)_{kn}  \\ 
 \left( {{\bf{F}}_{t,u}^c } \right)_{kn}  \leftarrow \left( {{\bf{F}}_{t,u}^c } \right)_{kn} \left( {{\bf{Sq}}_{t,u}^c } \right)_{kn}  \\ 
 \end{array} \right.
\end{equation}
where $\left( {{\bf{Sq}}_{t,l}^{v,s} } \right)_{kn}$,$\left( {{\bf{Sq}}_{t,l}^c } \right)_{kn}$,$\left( {{\bf{Sq}}_{t,u}^{v,s} } \right)_{kn}$,$\left( {{\bf{Sq}}_{t,u}^c } \right)_{kn}$ are defined as follows:

\begin{equation}
	\label{eq29}
\left\{ \begin{array}{l}
 \left( {{\bf{Sq}}_{t,l}^{v,s} } \right)_{kn}  = \sqrt {\frac{{\left[ {\pi _t^v \left( {{\bf{B}}_{t,s}^v } \right)^{\rm T} {\bf{X}}_{t,l}^v  + \beta \left( {H_ +  } \right)_t^{v,s} } \right]_{kn} }}{{\left[ {\pi _t^v \left( {{\bf{B}}_{t,s}^v } \right)^{\rm T} {\bf{B}}_t^v {\bf{F}}_{t,l}^v  + \beta \left( {H_ -  } \right)_t^{v,s} } \right]_{kn} }}}  \\ 
 \left( {{\bf{Sq}}_{t,l}^c } \right)_{kn}  = \sqrt {\frac{{\left[ {\sum\nolimits_{v = 1}^V {\pi _t^v \left( {{\bf{B}}_{t,c}^v } \right)^{\rm T} {\bf{X}}_{t,l}^v  + \beta \left( {H_ +  } \right)_t^c } } \right]_{kn} }}{{\left[ {\sum\nolimits_{v = 1}^V {\pi _t^v \left( {{\bf{B}}_{t,c}^v } \right)^{\rm T} {\bf{B}}_t^v {\bf{F}}_{t,l}^v  + \beta \left( {H_ -  } \right)_t^c } } \right]_{kn} }}}  \\ 
 \left( {{\bf{Sq}}_{t,u}^{v,s} } \right)_{kn}  = \sqrt {\frac{{\left[ {\pi _t^v \left( {{\bf{B}}_{t,s}^v } \right)^{\rm T} {\bf{X}}_{t,u}^v } \right]_{kn} }}{{\left[ {\pi _t^v \left( {{\bf{B}}_{t,s}^v } \right)^{\rm T} {\bf{B}}_t^v {\bf{F}}_{t,u}^v } \right]_{kn} }}}  \\ 
 \left( {{\bf{Sq}}_{t,u}^c } \right)_{kn}  = \sqrt {\frac{{\left[ {\sum\nolimits_{v = 1}^V {\pi _t^v \left( {{\bf{B}}_{t,c}^v } \right)^{\rm T} {\bf{X}}_{t,u}^v } } \right]_{kn} }}{{\left[ {\sum\nolimits_{v = 1}^V {\pi _t^v \left( {{\bf{B}}_{t,c}^v } \right)^{\rm T} {\bf{B}}_t^v {\bf{F}}_{t,u}^v } } \right]_{kn} }}}  \\ 
 \end{array} \right.
\end{equation}
where the matrices $\left( {{\bf{H}}_ +  } \right)_t $ and $\left( {{\bf{H}}_ -  } \right)_t $ are defined as follows:

\begin{equation}
	\label{eq30}
\begin{array}{l}
 \left( {{\bf{H}}_ +  } \right)_t  = \left( {\left( {{\bf{W}}_t  + {\bf{W}}_d } \right){\rm{ + }}\left( {{\bf{W}}_t  + {\bf{W}}_d } \right)^{\rm T} } \right)^ +  {\bf{F}}_{t,l}  + \left( {{\bf{W}}_t  + {\bf{W}}_d } \right)^ -  {\bf{Y}}_t  \\ 
 \;\;\;\;\;\;\;\; = [\left( {H_ +  } \right)_t^{1,s} ;\left( {H_ +  } \right)_t^{2,s} ; \cdots ;\left( {H_ +  } \right)_t^{V,s} ;\left( {H_ +  } \right)_t^c ] \\ 
 \left( {{\bf{H}}_ -  } \right)_t  = \left( {\left( {{\bf{W}}_t  + {\bf{W}}_d } \right){\rm{ + }}\left( {{\bf{W}}_t  + {\bf{W}}_d } \right)^{\rm T} } \right)^ -  {\bf{F}}_{t,l}  + \left( {{\bf{W}}_t  + {\bf{W}}_d } \right)^ +  {\bf{Y}}_t  \\ 
 \;\;\;\;\;\;\;\; = [\left( {H_ -  } \right)_t^{1,s} ;\left( {H_ -  } \right)_t^{2,s} ; \cdots ;\left( {H_ -  } \right)_t^{V,s} ;\left( {H_ -  } \right)_t^c ] \\ 
 \end{array}
\end{equation}

Let ${\bf{M}}^ +   = \left( {\left| {\bf{M}} \right| + {\bf{M}}} \right)/2$ and ${\bf{M}}^ -   = \left( {\left| {\bf{M}} \right| - {\bf{M}}} \right)/2$.

(6) Updating rule for ${\bf{\pi }}$:
The updating rule for ${\bf{\pi }}$ is similar to the formula (22). The overall AN-MTMVCSF algorithm is summarized in Algorithm 2.

\begin{algorithm}
	\renewcommand{\algorithmicrequire}{\textbf{Input:}}
	\renewcommand{\algorithmicensure}{\textbf{Output:}}
	\caption{}
	\label{alg2}
	\begin{algorithmic}[1]
		\REQUIRE  the instances sets of $V$ views in $T$ tasks: ${\bf{X}}_1^1 ,{\bf{X}}_1^2 , \cdots ,{\bf{X}}_1^V , \cdots ,{\bf{X}}_t^1 ,{\bf{X}}_t^2 , \cdots {\bf{X}}_t^V , \cdots ,{\bf{X}}_T^1 ,{\bf{X}}_T^2 , \cdots ,{\bf{X}}_T^V $; the label matrices of $T$ tasks: ${\bf{Y}}_1 ,{\bf{Y}}_2 , \cdots ,{\bf{Y}}_T$; the parameters: $\lambda$, $\beta$, $\gamma$,$\mu$.
		\ENSURE The feature representation matrices:${\bf{F}}_1 ,{\bf{F}}_2 , \cdots ,{\bf{F}}_T $; the weight matrices: ${\bf{W}}_1 ,{\bf{W}}_2 , \cdots ,{\bf{W}}_T $.\\
		\textbf{Initialization: } The basis matrix ${\bf{B}}_t^v $ of different views in each task; the feature matrix ${\bf{F}}_t^v $ of different views in each task; the weight matrices: ${\bf{W}}_1 ,{\bf{W}}_2 , \cdots ,{\bf{W}}_T $; the noise weight matrix $\bf{W}_d$; the positive semi-definite matrix ${\bf{D}}$; $\pi _t^v $ in vector ${\bf{\pi }}$ of different views in each tasks equal to $1/VT$.
		
		\textbf{Loop:}\\
		\textbf{For} each epoch \textbf{do:}\\
		\STATE Initialize ${\bf{B}}_t^v $,${\bf{F}}_t^v $,${\bf{W}}_1 ,{\bf{W}}_2 , \cdots ,{\bf{W}}_T $,$\bf{W}_d$,${\bf{D}}$,${\bf{\pi }}$;
		\STATE \textbf{For} each task $t \in \left\{ {1, \cdots ,T} \right\}$ \textbf{do}:\\
		\qquad Update ${\bf{W}}_t $ using (24);
		\STATE Update $\bf{W}_d$ using (25);
		\STATE Update ${\bf{D}}$ using (26);
		\STATE \textbf{For} each view $v \in \left\{ {1, \cdots ,V} \right\}$ \textbf{do}:\\
		\qquad \textbf{For} each task $t \in \left\{ {1, \cdots ,T} \right\}$ \textbf{do}:\\
		\qquad\qquad Update $\bf{B}_t^v$ using (27);\\
		\qquad\qquad Update ${\bf{F}}_{t,l}^{v,s} ,\;{\bf{F}}_{t,l}^c ,\;{\bf{F}}_{t,u}^{v,s} ,\;{\bf{F}}_{t,u}^c $ using (28);\\
		\qquad\qquad Update ${\bf{\pi }}$ using (22);
		\STATE ${\bf{F}}_t  \leftarrow \left[ {{\bf{F}}_{t,l}^{1,s} ,{\bf{F}}_{t,u}^{1,s} ; \cdots ;{\bf{F}}_{t,l}^{V,s} ,{\bf{F}}_{t,u}^{V,s} ;{\bf{F}}_{t,l}^c ,{\bf{F}}_{t,u}^c } \right]$
	\end{algorithmic}
\end{algorithm}

\section{Experiments and Discussion}

In this section, we design a series of experiments to demonstrate the effectiveness of MTMVCSF and AN-MTMVCSF on both real-world and synthetic data sets.

\subsection{Construction of Multi-task Multi-view Data Sets}

The first step of multi-task multi-view learning is to preprocess data because most of the data sets are not suitable to this typical learning scene directly. Although there is no mature multi-task multi-view data partitioning method nowadays, we will still give a short description about how we construct the data sets.

\textbf{WebKb}\footnote{ http://www.cs.cmu.edu/afs/cs/project/theo-20/www/data/} : The WebKb data set contains www-pages from computer science departments of four different universities (Cornell, Texas, Washington, Wisconsin), obviously, the four universities’ web information can be viewed as four tasks. And we can construct 3 views in each task: the words in each web page’s main text; the clickable words in each web page’s hyperlinks that pointing to other websites; and each website’s titles. For document preprocessing, we use Rainbow\footnote{http://www.cs.cmu.edu/~mccallum/bow/} , which can remove the header lines and stop words, and select words by mutual information. And note that although data set includes 7 categories, we only select 4 most representative ones to avoid category imbalance.

\textbf{20Newsgroup}\footnote{http://qwone.com/~jason/20Newsgroups/} : We call the 20 Newsgroups 20NG for short in this paper. The data set is organized into 20 different categories, and we sampled 200 documents from each newsgroup. At the same time, we define 20 tasks by transforming the multi-class problem into 20 binary classification problems. Obviously, for each task, we labeled the documents belong to the category as positive samples, and labeled the documents belong to the rest categories as negative samples. Note that we then sampled 200 negative samples to construct a task combined with this task’s positive samples to balance the categories. Furthermore, we constructed 21 views by viewing the words existing in all the tasks as a consistent view, and the words only appearing in corresponding task as a complementary view. Nevertheless, we find that only two views exist in all tasks, and finally these two views are utilized to construct the dataset \cite{12jin2013shared}. For document preprocessing, we first use the weighted tf-idf, and then use PCA \cite{45wold1987principal} to reduce each view’s features to 300.

\textbf{NUS-WIDE}: The Web Image Dataset contains 31 kinds of targets such as “fish”, “fox”, “cat”, “computer” and etc. We can define each of the categories as a task, that is to say, the database contains 31 tasks. However, some tasks have too few positive or negative examples, by removing them, a data set with only 6 tasks is obtained. Next, we download NUS-WIDE-OBJECT\footnote{http://lms.comp.nus.edu.sg/research/NUS-WIDE.htm}  which has 6 low-level features and we have selected five of them (all the features are included in download files): 64-dimensional color histogram, 144-dimensional color correlogram, 73-dimensional edge direction histogram, 128-dimensional wavelet texture, and 225-dimensional blockwise color moments. Each kind low-level feature is defined as a view, therefore the data set have five views in each task.

\textbf{Leaves}\footnote{https://archive.ics.uci.edu/ml/datasets/One-hundred+plant+species+leaves+data+set} : The one-hundred plant species leaves data set includes 100 species (or can be classified into 32 genera). We only selected the genus that include 3 or more than 3 species to construct 3 tasks. And in download files, they provide three types of features that can be viewed as three views together, including 64- dimensional shape descriptor, 64- dimensional fine scale margin and 64-dimensional texture histogram. As a result, the data set has three tasks with three views.

\textbf{Synth1}: This is a synthetic data set with 3 tasks and 5 views. We created 600 instances for each task and divided it into three equal parts. Each part is sampled from the normal distribution, but the combination of mean and standard deviation is (1, 1), (2, 2), (3, 3) respectively, and the corresponding label is 1, 2, 3. Furthermore, we use 5 filters to extract features, including averaging filter, maximum filter, Gaussian filter, approximates the two-dimensional Laplacian operator, and Prewitt horizontal edge-emphasizing filter. After the convolutional operations, we further map the values in data set between 0 and 1 using Min-Max Normalization.

\textbf{Synth2}: This is a synthetic data set with 4 tasks and 5 views. We created 800 instances for each task and divided it into four equal parts. Each part is sampled from the normal distribution, but the combination of mean and standard deviation is (1, 1), (2, 2), (3, 3), (4, 4) respectively, and the corresponding label is 1, 2, 3, 4. Furthermore, we use 5 filters to extract features, including averaging filter, maximum filter, Gaussian filter, approximates the two-dimensional Laplacian operator, and Prewitt horizontal edge-emphasizing filter. After the convolutional operations, we further map the values in data set between 0 and 1 using Min-Max Normalization.

For all data sets in our proposed methods, we should preprocess them in a simple way. For input data ${\bf{X}}_t^v  \in M_t^v  \times N$, we normalize each column and let the sum of all the elements in each column equals to 1. With this step, we can accelerate the convergency of algorithm and improve the learning performance. And for other baseline methods, the data fed to them does not need to be normalized, because they have their own preprocessing methods and we do not modify them. Besides, Webkb and NUS-WIDE are imbalanced data, where WebKb has 4 categories and the 4-th class data accounts for 56.64\% of whole data; NUS-WIDE has 2 categories and the second-class data accounts for 93.20\% of whole data. We solve this problem with proper over-sampling method, and according to the result shown in Table 5 and Table 7, MTMVCSF and RMTMVCSF perform better than the baseline methods we chosen in most cases. 

The properties of each dataset are listed in Table 2, where $V$, $T$, $C$, $Dim$ represent view numbers, task numbers, category numbers, and views’ feature dimensions in each task, respectively. 

\begin{table}[!htbp]
	\centering
	\label{tb2}
	\caption{Information on data sets.}
	\begin{tabular}{ccccc}
		\hline
		Data	& $V$	 & $T$	& $C$ 	& $Dim$\\\hline
		
WebKb&	3	&4	&4	&201-2500\\
20NewsGroup	&2	&20	&2	&300\\
NUS-WIDE	&6	&5	&2	&64-225\\
Leaves	&3	&3	&6	&64\\
Synth1	&5	&3	&3	&100\\
Synth2	&5	&4	&4	&100\\ \hline            
	\end{tabular}
\end{table}

\subsection{The selection of hyper-parameters}

Tradeoff parameters settings and dimension selection are regarded as important steps an effective training process, and all of them are hyper-parameters that must be determined before training commences. We define $K$ is the reconstructed dimension and define $KcPer$ is the percentage of common features, which means the percentage of special feature is $KsPer=1-KcPer$. In MTMVCSF, the hyper-parameters are $\beta$, $\gamma$, $K$ and $KcPer$, and in AN-MTMVCSF, in addition to $\beta$, $\gamma$, $K$ and $KcPer$, we also need to determine the value of hyper-parameter $\mu$. When we used grid method to find these parameters, the algorithm complexity is $O(n^4 )$ (MTMVCSF) or $O(n^5 )$ (AN-MTMVCSF) if we search on each parameter n times, which will cost a long time to find the best combination of parameters. However, we find that the choice of model parameters $\beta$, $\mu$, $\gamma$ has a little influence on the choice of dimension parameters $K$ and $KcPer$. Therefore, we separate the parameter tuning process, and the algorithm complexity is only $O(2n^2 )$ (MTMVCSF) or $O(n^3  + n^2 )$ (AN-MTMVCSF) to search the best configuration of parameters. Based on experience, the model parameters $\beta$, $\mu$, $\gamma$ are selected between $10^{-5}$ to 10, and $K$ is selected between 10 to 50. The configuration of hyper-parameters of MTMVCSF and AN-MTMVCSF are summarized in Table 3. When dealing with other data sets, we can just use WebKb’s hyperparameters configuration as shown in Table 3 first, and then fine-tune them with greedy method in their neighborhoods. If time permits, we’d better use segmented grid search method mentioned above. However, we also set the initial hyperparameters according to WebKb’s configuration and then fine-tune them in their neighborhoods.

\begin{table}[!htbp]
	\centering
	\label{tb3}
	\caption{Configuration of hyper-parameters.}
	\begin{tabular}{cccccccccc}
		\hline
\multirow{2}{*}{Data Set}       & \multicolumn{2}{c}{$\beta$} & $\mu$   &\multicolumn{2}{c}{$gamma$}   &\multicolumn{2}{c}{$K$}  & \multicolumn{2}{c}{$KcPer$}  \\\cline{2-10}
&Standard	&AN	&AN	&Standard	&AN	&Standard	&AN	&Standard	&AN\\ \hline
WebKb	     &$1e^{-4}$	&1$e^{-5}$	&$1e^{-3}$	&10	&0.1	&40	&20	&40\%	&60\%\\
20NewsGroup	&1$e^{-5}$	&1$e^{-5}$	&$1e^{-2}$	&1$e^{-5}$	&$1e^{-2}$	&20	&20	&60\%	&40\%\\
NUS-WIDE	&$1e^{-4}$	&1$e^{-5}$	&10	&1	&1	&20	&20	&60\%	&70\%\\
Leaves	&$1e^{-3}$	&$1e^{-3}$	&0.1	&$1e^{-2}$	&$1e^{-2}$	&20	&50	&60\%	&60\%\\
Synth1	&1$e^{-5}$	&1$e^{-5}$	&$1e^{-4}$	&$1e^{-4}$	&$1e^{-4}$	&50	&40	&40\%	&80\%\\
Synth2	&1$e^{-5}$	&$1e^{-4}$	&1	&$1e^{-2}$	&$1e^{-4}$	&50	&30	&40\%	&50\%\\\hline
	\end{tabular}
\end{table}

\begin{figure}[!htbp]
	\label{fig2}
	\centering
	\subfigure[Webkb]{
		\includegraphics[width=4.5cm]{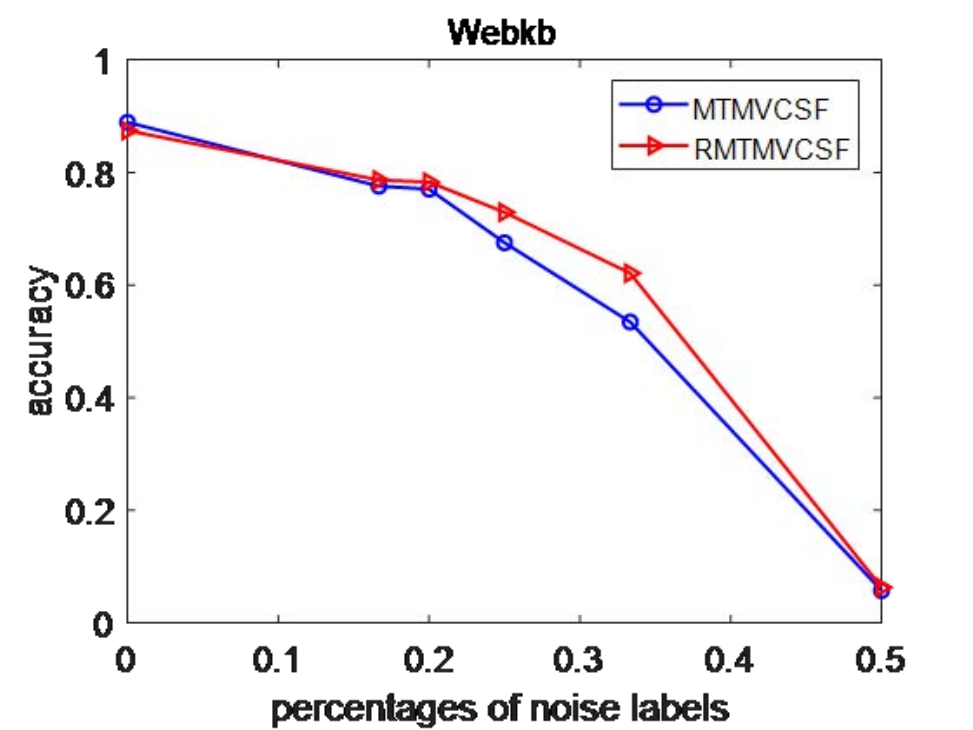}
	}
	\quad
	\subfigure[Leaves]{
		\includegraphics[width=4.5cm]{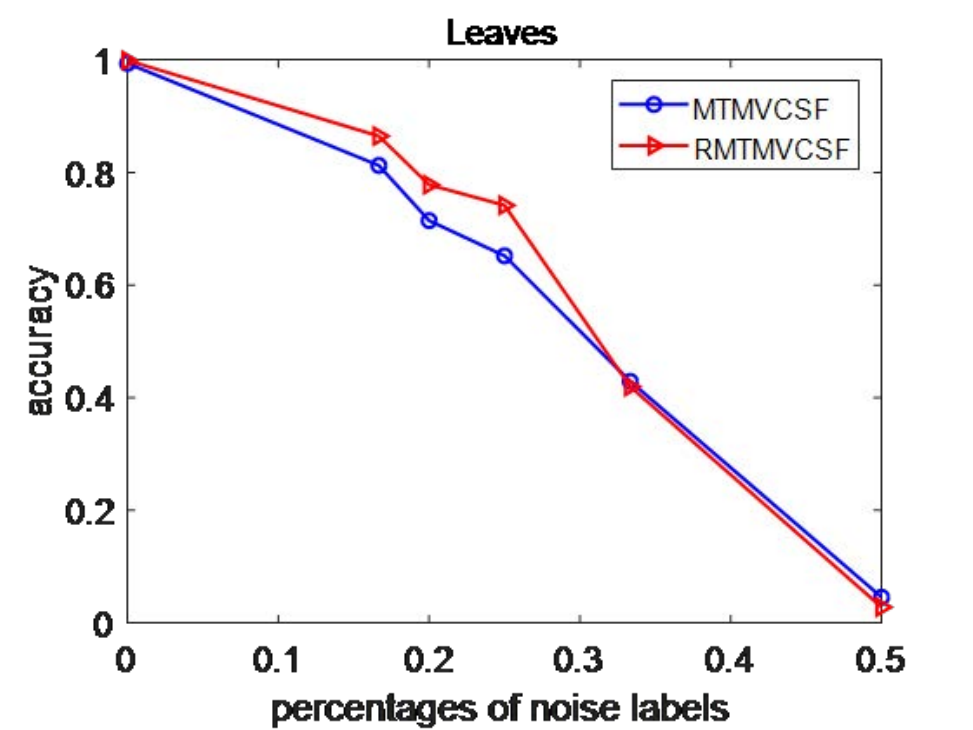}
	}
	\quad
	\subfigure[20NG]{
		\includegraphics[width=4.5cm]{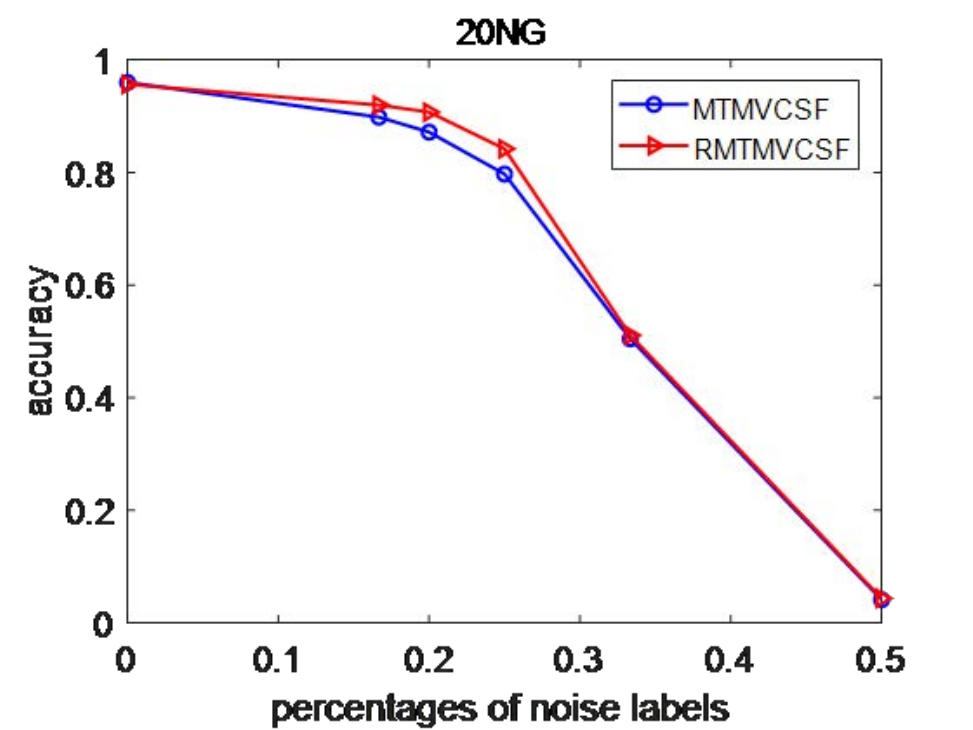}
	}
	\quad
	\subfigure[NUS-WIDE]{
		\includegraphics[width=4.5cm]{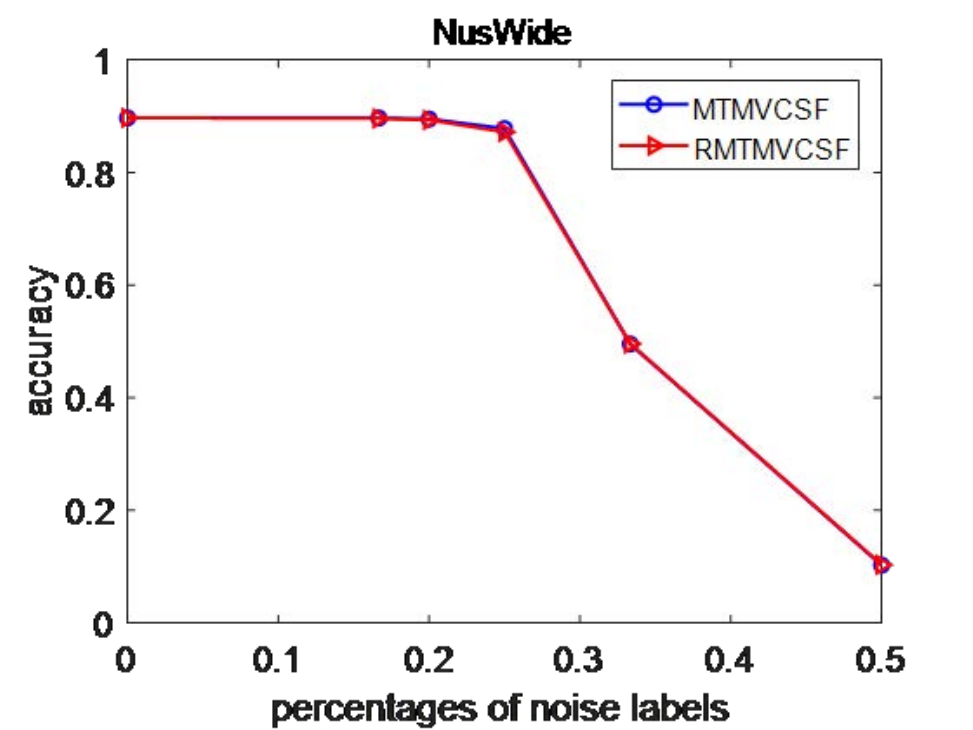}
	}
	\quad
	\subfigure[Synth1]{
		\includegraphics[width=4.5cm]{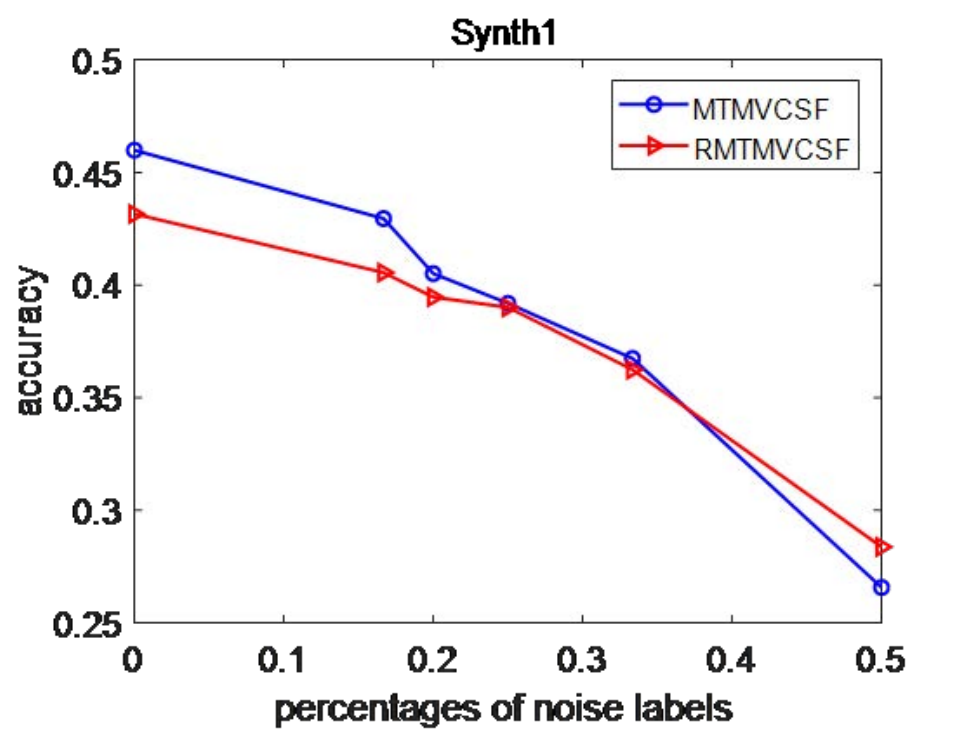}
	}
	\quad
	\subfigure[Synth2]{
		\includegraphics[width=4.5cm]{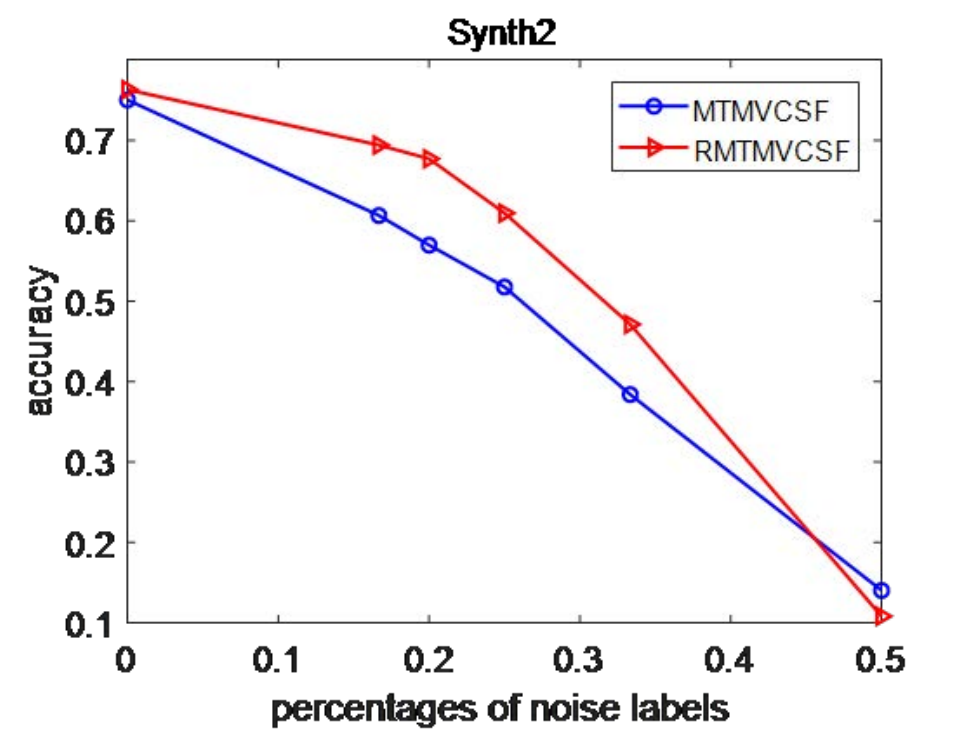}
	}
	\caption{Analysis of the ability of anti-noise.}
\end{figure}

\subsection{Anti-noise ability analysis of algorithm}

In real world application, the data sets always contain noise labels, which have harmful impact on the prediction performance. An anti-noise algorithm can avoid this kind of problem to a certain extent. Compared with MTMVCSF, AN-MTMVCSF has a better anti-noise ability. As we discussed previously in section 5, a noise weight is added in training stage, thus, AN-MTMVCSF is more adaptable to the noise labels.

In this subsection, we will add noise to labels and observe the performance of MTMVCSF and AN-MTMVCSF in a classification task. We use ${\bf{W}}_t {\bf{F}}_{t,u}$ to predict the unlabeled labels of task t, where ${\bf{W}}_t $ is the model parameters learned by MTMVCSF and AN-MTMVCSF, and ${\bf{F}}_{t,u}$ is the latent feature learned according to MTMVCSF and AN-MTMVCSF. The proportion of noise will increase gradually from 0\% to 50\%. With the increase of noise, the accuracy of different algorithms changes as shown in Fig. 2. Obviously, we can easily find that when labels have no noise, the performance of MTMVCSF will be better, especially in the data set of Synth1. However, with the increase of noise, the MTMVCSF’s performance drops significantly faster than AN-MTMVCSF’s, and the classification accuracy of AN-MTMVCSF performs much better with the noise labels in both real and synthetic data sets.
Consequently, AN-MTMVCSF has a better anti-noise ability than MTMVCSF, and it can effectively avoid the influence caused by noise labels.

\subsection{Convergence analysis of the algorithm}

The convergence of an algorithm and its convergence rate can be used to evaluate the efficiency the algorithm. With the increase of the iteration numbers, the convergence of the algorithm can be judged by the value of loss function. To make comparison fair, we normalized the values of the loss functions under different data sets, and Fig. 3 and Fig. 4 give an intuitive illustration. Obviously, the values of loss function converge well both in MTMVCSF and AN-MTMVCSF, and we can easily find out that the algorithm converges in 20-30 iterations on four data sets, which is really a wonderful result. Besides, we recorded the running time of the algorithm using Matlab 2017b on a standard Window PC with an Intel 3.00-GHz CPU and 16-GB RAM. For all data sets, we conduct experiment using MTMVCSF and AN-MTMVCSF for 50 iterations respectively, and record the running times in seconds. The results are summarized in Table 4.

\begin{figure}[!htbp]
	\centering
	\includegraphics[scale=1]{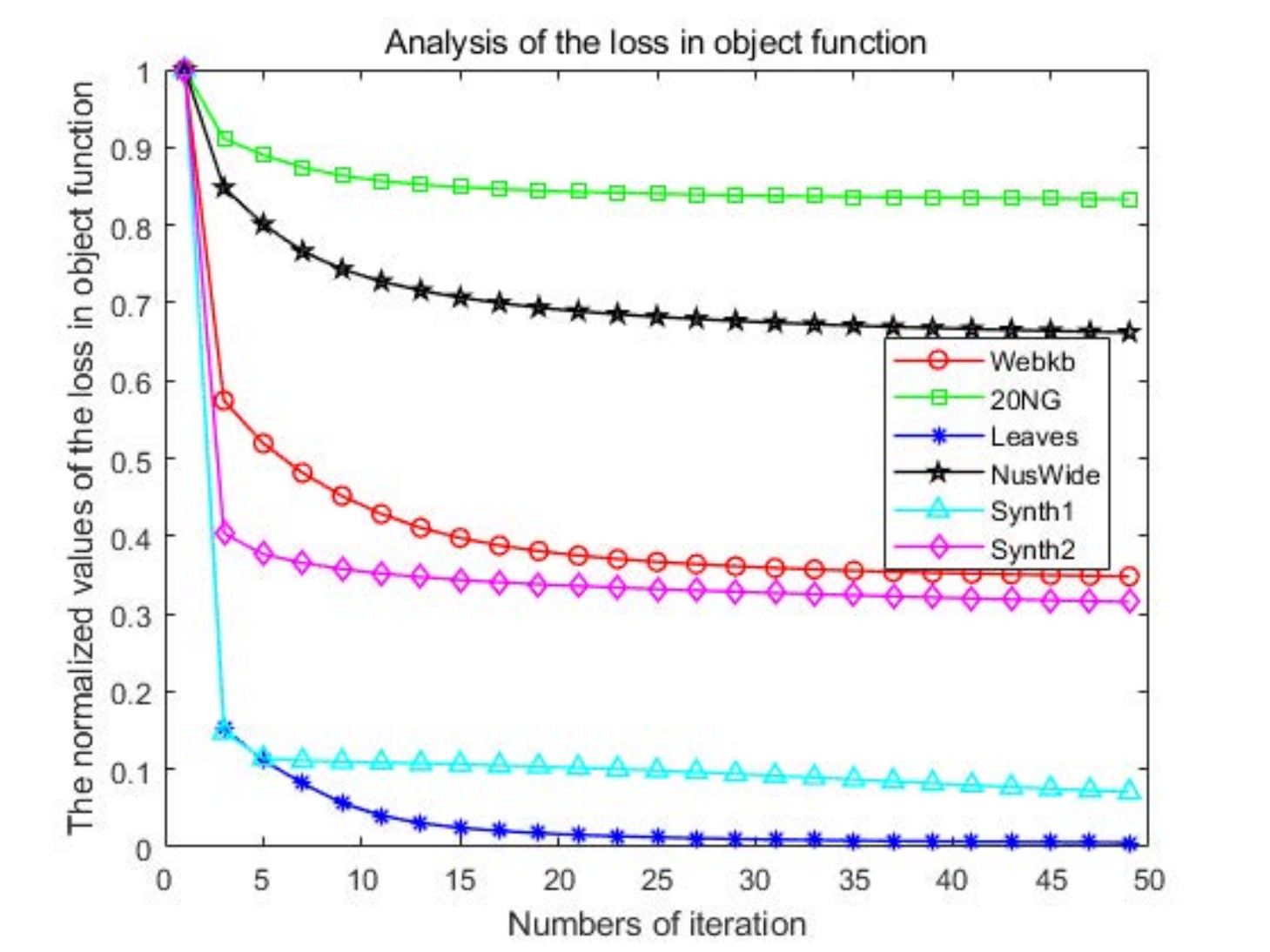}
	\caption{Loss of MTMVCSF.}
	\label{fig3}
\end{figure}
\begin{figure}[!htbp]
	\centering
	\includegraphics[scale=1]{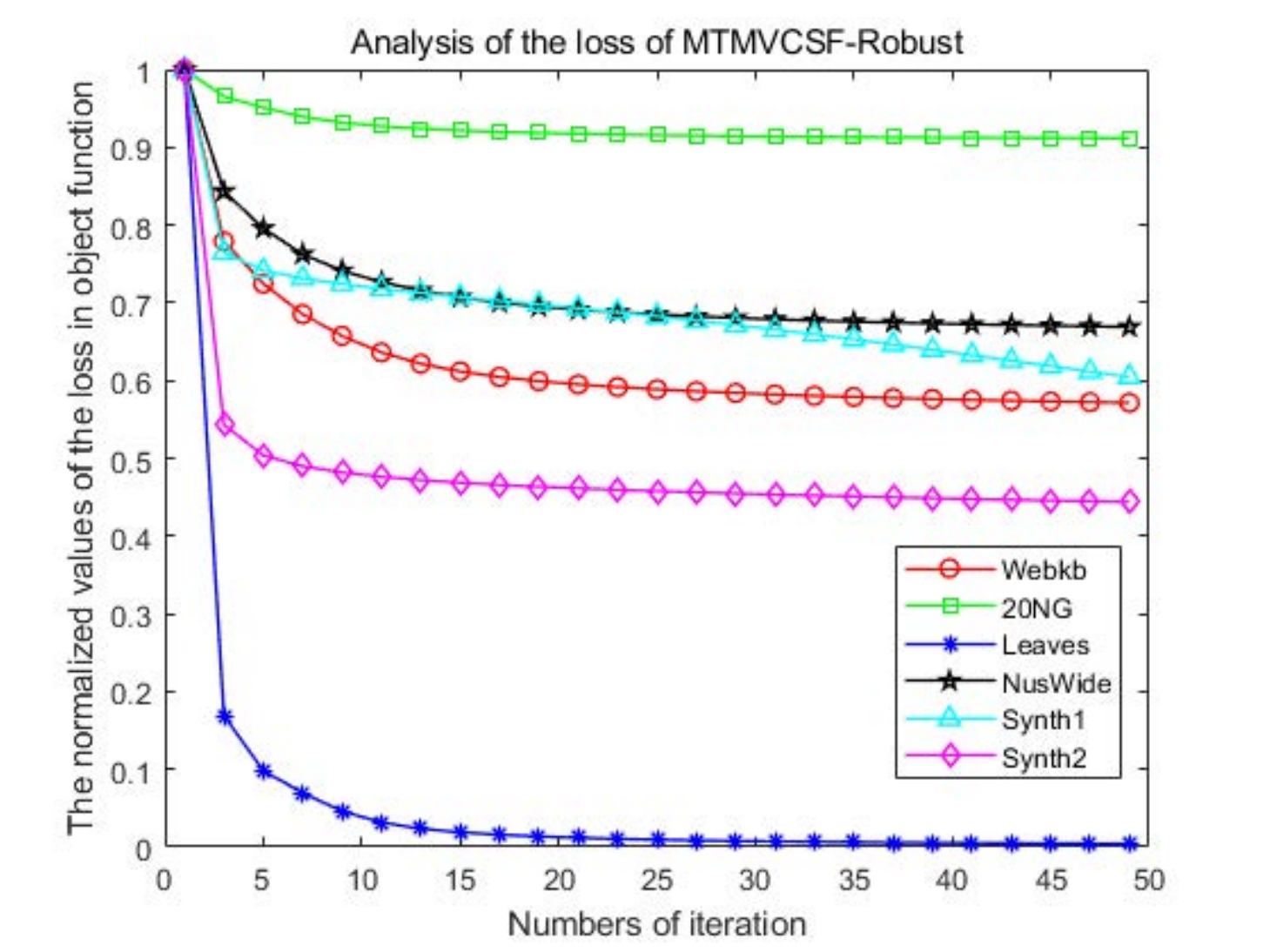}
	\caption{Loss of AN-MTMVCSF.}
	\label{fig4}
\end{figure}

\begin{table}[!htbp]
	\centering
	\caption{Running Times (50 iterations)}
	\begin{tabular}{ccccccc}
		\hline
		Method & \multicolumn{6}{c}{MTMVCSF}\\\hline
		Data Set	&Leaves	&20NG	&Webkb	&NusWide	&Synth1	&Synth2\\
Times (s)	&0.99	&3.33	&3.28	&19.97	&1.97	&3.52\\\hline
		Method & \multicolumn{6}{c}{AN-MTMVCSF}\\\hline
		Data Set	&Leaves	&20NG	&Webkb	&NusWide	&Synth1	&Synth2\\
Times (s)	&1.47	&2.48	&2.21	&19.83	&1.20	&2.32\\\hline
	\end{tabular}
\end{table}

\subsection{Classification and Clustering performance}

In this paper, the fundamental purpose of our proposed method is to find each task’s latent feature ${\bf{F}}_{t,u}$ corresponding to those unlabeled samples. Although we also obtained each task’s labeled latent feature ${\bf{F}}_{t,l}$, its major role is to supervise the division of matrix factorization and help unlabeled samples find better latent representations. As a result, ${\bf{F}}_{t,u}$ can be viewed as the substitution of input sample of task $t$, afterwards it can be used effectively for some learning tasks, such as classification and clustering. Concurrently to our work, PSLF \cite{5liu2014partially} has also proposed another feature representation learning method which obtain a latent feature composed of special feature in each view and common feature of all views, however, they do not consider the influence of multiple tasks. As a consequence, we can obtain the latent feature of raw data from three aspects which involve PSLF, MTMVCSF, AN-MTMVCSF, we will use the learned latent representation to conduct some experiments to compare and verify the performance of these methods.

\begin{table}[]
	\centering
	\caption{Classification results}
	\resizebox{\textwidth}{110mm}{
	\begin{tabular}{cccccc}
		\hline
\multicolumn{2}{c}{}  & LR &LR-PSLF	&LR-MTMVCSF &LR-AN-MTMVCSF \\ \hline
\multirow{5}{*}{Leaves}&Accuracy	&90.05\%$\pm$6.572\%&98.26\%$\pm$ 1.870\%
&100\%$\pm$ 0\%
&100\%$\pm$0\%\\
&Precision	&90.99\%$\pm$ 7.118\%&98.53\%$\pm$ 1.542\%&100\%$\pm$ 0\%&100\%$\pm$ 0\%\\
&F1	&89.45\%$\pm$ 7.154\%&98.27\%$\pm$ 1.867\%&100\%$\pm$ 0\%&100\%$\pm$ 0\%\\
&DICE	&0.8277$\pm$ 0.3776&0.8277$\pm$ 0.3790&0.8267$\pm$ 0.3791&0.8261$\pm$ 0.3790\\
&Jaccard	&90.05\%$\pm$ 6.572\%&98.26\%$\pm$ 1.870\%&100\%$\pm$ 0\%&100\%$\pm$ 0\%\\\hline
\multirow{5}{*}{20NG}	&Accuracy	&97.21\%$\pm$ 2.431\%&95.10\%$\pm$ 4.342\%&96.40\%$\pm$ 4.0615\%&95.90\%$\pm$ 4.0899\%\\
&Precision	&97.33\%$\pm$ 2.228\%&95.23\%$\pm$ 4.201\%&96.52\%$\pm$ 4.0535\%&95.95\%$\pm$ 4.0889\%\\
&F1	&97.20\%$\pm$ 2.441\%&95.08\%$\pm$ 4.351\%&96.39\%$\pm$ 4.0624\%&95.89\%$\pm$ 4.0904\%\\
&DICE	&0.4983$\pm$ 0.4999&0.4983$\pm$ 0.5000&0.4981$\pm$ 0.5000&0.4983$\pm$ 0.5000\\
&Jaccard	&97.21\%$\pm$ 2.431\%&95.10\%$\pm$ 4.342\%&96.40\%$\pm$ 4.0615\%&95.90\%$\pm$ 4.0899\%\\\hline
\multirow{5}{*}{Webkb}
&Accuracy	&79.60\%$\pm$  6.323\%
&86.37\%$\pm$  3.096\%
&88.72\%$\pm$  2.1111\%
&87.17\%$\pm$  3.0098\%\\
&Precision	&80.22\%$\pm$  8.474\%
&86.95\%$\pm$  3.121\%
&89.48\%$\pm$  3.1128\%
&87.83\%$\pm$  3.0238\%\\
&F1	&67.31\%$\pm$  10.05\%
&81.39\%$\pm$  3.943\%
&85.90\%$\pm$  3.1757\%
&84.17\%$\pm$  3.0153\%\\
&DICE	&0.6096$\pm$  0.4878
&0.6427$\pm$  0.4792
&0.6429$\pm$  0.4791
&0.6337$\pm$  0.4792\\
&Jaccard	&79.60\%$\pm$  6.323\%
&86.37\%$\pm$  3.096\%
&88.72\%$\pm$  2.1111\%
&87.17\%$\pm$  3.0098\%\\\hline
\multirow{5}{*}{NusWide}
&Accuracy	&89.70\%$\pm$ 3.069\%
&89.85\%$\pm$ 2.949\%
&89.67\%$\pm$ 2.0877\%
&89.77\%$\pm$ 2.0771\%\\
&Precision	&81.61\%$\pm$ 4.908\%
&82.59\%$\pm$ 4.186\%
&83.81\%$\pm$ 4.0694\%
&83.66\%$\pm$ 4.0697\%\\
&F1	&48.42\%$\pm$ 3.353\%
&48.85\%$\pm$ 3.005\%
&49.82\%$\pm$ 3.2729\%
&50.47\%$\pm$ 3.4053\%\\
&DICE	&0.1064$\pm$ 0.0442
&0.1058$\pm$ 0.3076
&0.1034$\pm$ 0.3045
&0.1039$\pm$ 0.3045\\
&Jaccard	&89.70\%$\pm$ 3.069\%
&89.85\%$\pm$ 2.949\%
&89.67\%$\pm$ 2.0877\%
&89.77\%$\pm$ 2.0771\%\\\hline
\multicolumn{2}{c}{} &RF	&RF-PSLF	&RF-MTMVCSF	&RF-AN-MTMVCSF\\\hline
\multirow{5}{*}{Leaves} 
&Accuracy	&88.66\%$\pm$ 5.983\%
&97.57\%$\pm$ 2.528\%
&99.31\%$\pm$ 2.0096\%
&100\%$\pm$ 0\%\\
&Precision	&90.76\%$\pm$ 5.688\%
&98.06\%$\pm$ 1.89\%
&99.39\%$\pm$ 1.0074\%
&100\%$\pm$ 0\%\\
&F1	&88.52\%$\pm$ 6.206\%
&97.65\%$\pm$ 2.385\%
&99.30\%$\pm$ 2.0097\%
&100\%$\pm$ 0\%\\
&DICE	&0.8264$\pm$ 0.3788
&0.8262$\pm$ 0.3789
&0.8261$\pm$ 0.3790
&0.8256$\pm$ 0.3794\\
&Jaccard	&88.66\%$\pm$ 5.983\%
&97.57\%$\pm$ 2.528\%
&99.31\%$\pm$ 2.0096\%
&100\%$\pm$ 0\%\\\hline
\multirow{5}{*}{20NG}
&Accuracy	&95.56\%$\pm$ 3.222\%
&96.04\%$\pm$ 3.549\%
&95.58\%$\pm$ 2.1224\%
&96.47\%$\pm$ 2.524\%\\
&Precision	&95.77\%$\pm$ 2.959\%
&96.14\%$\pm$ 0.1191\%
&95.96\%$\pm$ 2.717\%
&96.57\%$\pm$ 2.402\%\\
&F1	&95.54\%$\pm$ 3.238\%
&96.02\%$\pm$ 3.557\%
&95.83\%$\pm$ 2.821\%
&96.46\%$\pm$ 2.538\%\\
&DICE	&0.4987$\pm$ 0.5000
&0.4981$\pm$ 0.5000
&0.4984$\pm$ 0.5000
&0.4982$\pm$ 0.5000\\
&Jaccard	&95.56\%$\pm$ 3.222\%
&96.04\%$\pm$ 3.549\%
&95.58\%$\pm$ 2.1224\%
&96.47\%$\pm$ 2.524\%\\\hline
\multirow{5}{*}{Webkb}
&Accuracy	&76.42\%$\pm$ 6.560\%
&85.02\%$\pm$ 4.264\%
&85.42\%$\pm$ 4.352\%
&84.60\%$\pm$ 4.899\%\\
&Precision	&76.25\%$\pm$ 9.964\%
&86.40\%$\pm$ 3.524\%
&86.73\%$\pm$ 3.661\%
&85.39\%$\pm$ 5.822\%\\
&F1	&59.27\%$\pm$ 11.55\%
&78.44\%$\pm$ 5.385\%
&79.22\%$\pm$ 5.466\%
&76.91\%$\pm$ 7.763\%\\
&DICE	&0.5866$\pm$ 0.4924
&0.6236$\pm$ 0.4845
&0.6213$\pm$ 0.4851
&0.6220$\pm$ 0.4849\\
&Jaccard	&76.42\%$\pm$ 6.560\%
&85.02\%$\pm$ 4.264\%
&85.42\%$\pm$ 4.362\%
&84.60\%$\pm$ 4.899\%\\\hline
\multirow{5}{*}{NusWide}
&Accuracy	&89.65\%$\pm$ 3.108\%
&89.66\%$\pm$ 0.3210\%
&89.89\%$\pm$ 2.771\%
&89.64\%$\pm$ 3.272\%\\
&Precision	&81.48\%$\pm$ 4.978\%
&80.50\%$\pm$ 5.715\%
&83.96\%$\pm$ 4.521\%
&80.44\%$\pm$ 5.823\%\\
&F1	&47.84\%$\pm$ 2.056\%
&47.26\%$\pm$ 0.8987\%
&50.53\%$\pm$ 6.235\%
&47.25\%$\pm$ 0.9169\%\\
&DICE	&0.1051$\pm$ 0.0422
&0.1034$\pm$ 0.3044
&0.1112$\pm$ 0.3144
&0.1037$\pm$ 0.3049\\
&Jaccard	&89.65\%$\pm$ 3.108\%
&89.66\%$\pm$ 0.3210\%
&89.89\%$\pm$ 2.771\%
&89.63\%$\pm$ 3.272\%\\\hline
	\end{tabular}}
\end{table}

\textbf{Classification:} Two baseline classification methods, i.e. Logistic Regression (LR) and Random Forest (RF) are selected, and we feed different input samples come from raw data, PSLF, MTMVCSF and AN-MTMVCSF to LR and RF, we call the different methods with different inputs as LR, LR-PSLF, LR-MTMVCSF, LR-AN-MTMVCSF and RF, RF-PSLF, RF-MTMVCSF, RF-AN-MTMVCSF. The evaluation metrics are accuracy, precision, F1, DICE, and Jaccard. On the other hand, we random split the data by 50\% into training and testing data. The overall results are listed in Table 5, and obviously we find that the classification performance gets better when using latent representation, especially LR-MTMVCSF, LR-AN-MTMVCSF, RF-MTMVCSF, and RF-AN-MTMVCSF perform great. 

Beside this, we also choose some classification methods as baselines to compare with the results in Table 5, and they all can handle multi-task multi-view problems: 

\textbf{$IteM^2$:} $IteM^2$ \cite{9he2011graphbased} is a graph-based iterative algorithm, which can take full advantage use of the dual-heterogeneity; 

\textbf{CSL\_MTMV:} CSL\_MTMV \cite{12jin2013shared} is convex multi-task multi-view problem considering shared structure using induction. The algorithm can learn shared predictive structures on related views through multiple correlative tasks, and use the common information among different views;

\textbf{MAMUDA:} MAMUDA \cite{13jin2014multi} is a multi-task multi-view discriminant analysis algorithm. By exploring the inherent structure of shared task-specific and problem-specific, it collaboratively learns the feature transforms of different views and tasks.

The experimental results are listed in Table 6.

\begin{table}[!htbp]
	\centering
	\caption{Baseline classification methods}
	\begin{tabular}{cccccc}
		\hline
		\multicolumn{2}{c}{} & $IteM^2$	&CSL\_MTMV	&MAMUDA\\\hline
\multirow{5}{*}{Leaves}
&Accuracy	&94.56\%	&99.32\%	&100\%\\
&Precision	&95.28\%	&98.99\%	&100\%\\
&F1	&95.43\%	&98.04\%	&100\%\\
&DICE	&0.8317	&0.8297	&0.8288\\
&Jaccard	&94.56\%	&99.32\%	&100\%\\\hline
\multirow{5}{*}{20NG}
&Accuracy	&65.66\%	&97.63\%	&85.87\%\\
&Precision	&64.28\%	&98.90\%	&86.11\%\\
&F1	     & 65.47\%	&98.73\%	&86.11\%\\
&DICE	     & 0.4991	&0.4981	&0.4982\\
&Jaccard	& 65.66\%	&97.63\%	&85.87\%\\\hline
\multirow{5}{*}{Webkb}
&Accuracy	&76.22\%	&80.34\%	&86.23\%\\
&Precision	&75.34\%	&81.32\%	&88.25\%\\
&F1	     &73.66\%	&81.26\%	&81.03\%\\
&DICE	     &0.6429	&0.6389	&0.6401\\
&Jaccard	&76.22\%	&80.34\%	&86.23\%\\\hline
\multirow{5}{*}{Webkb}
&Accuracy	&59.46\%	&64.31\%	&82.48\%\\
&Precision	&19.65\%	&20.03\%	&83.01\%\\
&F1	     & 12.68\%	&13.27\%	&83.23\%\\
&DICE	     & 0.1109	&0.1103	&0.1092\\
&Jaccard	&59.46\%	&64.31\%	&82.48\%\\\hline
	\end{tabular}
\end{table}

\textbf{Semi-Clustering}: Two baseline clustering methods Gaussian Mixture Model (GMM) and Affinity Propagation (AP) are selected. A semi-supervised clustering pattern is adopted in this experiment, we utilize a small part of data set to help the remaining data find the latent representation with the methods of PSLF, MTMVCSF, and AN-MTMVCSF respectively, and then we clustering the latent features. Specifically, we choose 15\% data to train latent feature with the methods of PSLF, MTMVCSF, and AN-MTMVCSF respectively, and the remaining data is used to clustering. We call the different methods as GMM, GMM-PSLF, GMM-MTMVCSF, GMM-AN-MTMVCSF and AP, AP-PSLF, AP-MTMVCSF, AP-AN-MTMVCSF. The evaluation metrics are Normalized Mutual Information (NMI), Adjusted Rand Index (ARI), Homogeneity, Completeness. The overall results are listed in Table 7, obviously the effect of clustering has been improved significantly on four real-world data sets.

Beside this, we also choose some classification methods as baselines to compare with the results in Table 7, and they all can handle multi-task multi-view problems. Bipartite graph based Multi-Task Multi-View Clustering (BMTMVC) and Semi-nonnegative matrix tri-factorization based Multi-Task Multi-View Clustering (SMTMVC) are all proposed in \cite{1zhang2016multi}, BMTMVC can be only adopted when facing nonnegative data, but SMTMVC is a general method that can deal with negative data. The overall results are listed in Table 8.

Note that in the experiments above, we use ${\bf{F}}_{t,u}$ and ${\bf{X}}_{{\rm{t,u}}}^{\rm{v}}$ in classification and clustering tasks, where ${\bf{X}}_{{\rm{t,u}}}^{\rm{v}}$ is the original data of ${\bf{F}}_{t,u}$. For classification, we split both ${\bf{F}}_{t,u}$ and ${\bf{X}}_{{\rm{t,u}}}^{\rm{v}}$ by 50\%, and ${\bf{X}}_{{\rm{t,u}}}^{\rm{v}}$ is the data set for LR and RF, the latent features are for LRPSLF, LR\_MTMVCSF, LR\_AN-MTMVCSF, RF\_PSLF, RF\_MTMVCSF, and RF\_AN-MTMVCSF. For clustering task, we conduct clustering on ${\bf{F}}_{t,u}$ and ${\bf{X}}_{{\rm{t,u}}}^{\rm{v}}$ directly, and ${\bf{X}}_{{\rm{t,u}}}^{\rm{v}}$ is the data set for Gaussian Mixture Model (GMM) and Affinity Propagation (AP), the latent features are for GMM\_PSLF, GMM\_MTMVCSF, GMM\_AN-MTMVCSF, AP\_PSLF, AP\_MTMVCSF, and AP\_AN-MTMVCSF.

\begin{table}[!htbp]
	\centering
	\caption{Clustering results}
	\resizebox{\textwidth}{90mm}{
	\begin{tabular}{cccccc}
		\hline
		\multicolumn{2}{c}{} & Gaussian	&Gaussian-PSLF	&Gaussian-MTMVCSF	&Gaussian-AN-MTMVCSF\\\hline
		\multirow{4}{*}{Leaves} 
&NMI	&53.76\%$\pm$ 14.83\%
&91.39\%$\pm$ 0.1065\%
&90.58\%$\pm$ 0.1437\%
&95.61\%$\pm$ 0.1386\%\\
&ARI	&45.71\% 10.86\%
&84.45\%$\pm$ 0.5457\%
&82.84\%$\pm$ 0.6424\%
&90.74\%$\pm$ 0.8448\%\\
&Homogeneity	&52.76\%$\pm$ 14.26\%
&90.04\%$\pm$ 0.1943\%
&88.80\%$\pm$ 0.2752\%
&94.44\%$\pm$ 0.2742\%\\
&Completeness	&88.13\%$\pm$ 1.147\%
&92.79\%$\pm$ 0.0595\%
&92.45\%$\pm$ 0.0929\%
&96.83\%$\pm$ 0.0538\%\\		
		\hline
		\multirow{4}{*}{20NG}
&NMI	&28.22\%$\pm$ 6.336\%
&36.07\%$\pm$ 6.160\%
&38.25\%$\pm$ 5.500\%
&37.38\%$\pm$ 7.476\%\\
&ARI	&23.64\%$\pm$ 8.629\%
&38.81\%$\pm$ 8.277\%
&36.21\%$\pm$ 8.979\%
&37.77\%$\pm$ 11.59\%\\
&Homogeneity	&25.73\%$\pm$ 6.795\%
&35.15\%$\pm$ 6.409\%
&35.97\%$\pm$ 6.189\%
&35.63\%$\pm$ 8.264\%\\
&Completeness	&32.17\%$\pm$ 5.490\%
&37.49\%$\pm$ 5.683\%
&41.30\%$\pm$ 4.614\%
&40.07\%$\pm$ 6.369\%	\\	
		 \hline
		\multirow{4}{*}{Webkb}     
&NMI	&24.81\%$\pm$ 4.448\%
&31.48\%$\pm$ 1.849\%
&34.77\%$\pm$ 0.2706\%
&36.37\%$\pm$ 0.3825\%\\
&ARI	&15.19\%$\pm$ 4.750\%
&29.97\%$\pm$ 2.592\%
&39.55\%$\pm$ 0.3790\%
&35.67\%$\pm$ 0.7078\%\\
&Homogeneity	&20.20\%$\pm$ 4.776\%
&32.77\%$\pm$ 1.896\%
&35.60\%$\pm$ 0.2039\%
&36.83\%$\pm$ 0.3585\%\\
&Completeness	&33.69\%$\pm$ 3.935\%
&30.27\%$\pm$ 1.811\%
&33.98\%$\pm$ 0.3447\%
&35.98\%$\pm$ 0.4604\%\\		
		\hline
		\multirow{4}{*}{NusWide}
&NMI	&0.544\%$\pm$ 0.0178\%
&0.669\%$\pm$ 0.0171\%
&0.962\%$\pm$ 0.0343\%
&1.132\%$\pm$ 0.0553\%\\
&ARI	&0.549\%$\pm$ 0.0738\%
&0.569\%$\pm$ 0.0200\%
&0.516\%$\pm$ 0.0157\%
&0.896\%$\pm$ 0.0482\%\\
&Homogeneity	&0.688\%$\pm$ 0.0273\%
&0.878\%$\pm$ 0.0272\%
&1.265\%$\pm$ 0.0548\%
&1.454\%$\pm$ 0.0884\%\\
&Completeness	&0.433\%$\pm$ 0.0116\%
&0.515\%$\pm$ 0.0107\%
&0.736\%$\pm$ 0.0214\%
&0.883\%$\pm$ 0.0346\%\\\hline
\multicolumn{2}{c}{} &AP	&AP-PSLF	&AP-MTMVCSF	&AP-AN-MTMVCSF\\\hline
		\multirow{4}{*}{Leaves} 
&NMI	&80.68\%$\pm$ 0.6751\%
&91.43\%$\pm$ 0.0252\%
&91.77\%$\pm$ 0.0356\%
&95.08\%$\pm$ 0.0327\%\\
&ARI	&68.44\%$\pm$ 0.9185\%
&85.97\%$\pm$ 0.1694\%
&87.07\%$\pm$ 0.0122\%
&93.13\%$\pm$ 0.8448\%\\
&Homogeneity	&81.33\%$\pm$ 0.3576\%
&92.53\%$\pm$ 0.0112\%
&92.40\%$\pm$ 0.1323\%
&96.65\%$\pm$ 0.0132\%\\
&Completeness	&80.30\%$\pm$ 1.434\%
&90.38\%$\pm$ 0.1075\%
&91.18\%$\pm$ 0.0036\%
&93.55\%$\pm$ 0.0799\%	\\	
		\hline
		\multirow{4}{*}{20NG} 
&NMI	&33.03\%$\pm$ 0.2493\%
&35.36\%$\pm$ 0.3560\%
&35.64\%$\pm$ 1.1249\%
&35.37\%$\pm$ 0.9317\%\\
&ARI	&5.358\%$\pm$ 0.0802\%
&21.14\%$\pm$ 0.7493\%
&24.96\%$\pm$ 1.5250\%
&25.22\%$\pm$ 1.0927\%\\
&Homogeneity	&73.21\%$\pm$ 1.449\%
&56.72\%$\pm$ 2.4552\%
&50.13\%$\pm$ 2.6070\%
&48.79\%$\pm$ 2.2717\%\\
&Completeness	&14.94\%$\pm$ 0.0517\%
&22.12\%$\pm$ 0.3880\%
&25.49\%$\pm$ 0.5665\%
&25.79\%$\pm$ 0.4147\%	\\	
		\hline
		\multirow{4}{*}{Webkb} 
&NMI	&35.78\%$\pm$ 3.8511\%
&36.81\%$\pm$ 0.2807\%
&37.24\%$\pm$ 0.1242\%
&36.68\%$\pm$ 0.3190\%\\
&ARI	&13.38\%$\pm$ 4.6880\%
&12.11\%$\pm$ 0.0576\%
&14.26\%$\pm$ 0.2236\%
&13.48\%$\pm$ 0.0875\%\\
&Homogeneity	&52.57\%$\pm$ 14.551\%
&54.44\%$\pm$ 0.7173\%
&54.57\%$\pm$ 0.3604\%
&54.34\%$\pm$ 0.6987\%\\
&Completeness	&37.32\%$\pm$ 6.0226\%
&24.91\%$\pm$ 0.1120\%
&25.44\%$\pm$ 0.0494\%
&24.76\%$\pm$ 0.1463\%	\\
		\hline
		\multirow{4}{*}{NusWide} 
&NMI	&3.398\%$\pm$ 0.0313\%
&3.691\%$\pm$ 0.0226\%
&4.532\%$\pm$ 0.0321\%
&4.746\%$\pm$ 0.0364\%\\
&ARI	&0.012\%$\pm$ 0.0004\%
&0.009\%$\pm$ 0.0001\%
&0.016\%$\pm$ 0.0002\%
&0.016\%$\pm$ 0.0001\%\\
&Homogeneity	&13.39\%$\pm$ 0.3579\%
&15.09\%$\pm$ 0.3122\%
&18.94\%$\pm$ 0.4874\%
&19.93\%$\pm$ 0.5264\%\\
&Completeness	&0.887\%$\pm$ 0.0029\%
&0.916\%$\pm$ 0.0020\%
&1.100\%$\pm$ 0.0022\%
&1.146\%$\pm$ 0.0001\%	\\	
		\hline
	\end{tabular}}
\end{table}

\begin{table}[!htbp]
	\centering
	\caption{Baseline clustering methods}
	\resizebox{\textwidth}{15mm}{
	\begin{tabular}{ccccccccc}
		\hline
\qquad &\multicolumn{2}{c}{Leaves}	&\multicolumn{2}{c}{Webkb}	&\multicolumn{2}{c}{20NG}	&\multicolumn{2}{c}{NusWide}\\\cline{2-9}
&SMTMVC	&BMTMVC	&SMTMVC	&BMTMVC	&SMTMVC	&BMTMVC	&SMTMVC	&BMTMVC\\\hline	
NMI	&89.82\%	&89.64\%	&21.82\%	&44.81\%	&34.35\%	&34.25\%	&3.655\%	&2.854\%\\
ARI	&88.42\%	&86.95\%	&25.17\%	&55.02\%	&5.962\%	&6.232\%	&0.012\%	&0.016\%\\
Homogeneity	&89.68\%	&89.28\%	&19.80\%	&41.46\%	&73.55\%	&74.23\%	&12.29\%	&13.02\%\\
Completeness	&89.96\%	&89.99\%	&24.13\%	&48.44\%	&15.43\%	&14.28\%	&0.855\%	&0.745\%\\
\hline
	\end{tabular}}
\end{table}

\subsection{Combination with the Baseline Methods}

PSLF, MTMVCSF, and AN-MTMVCSF are all the methods to find the latent features of raw data, compared with PSLF, not only MTMVCSF and AN-MTMVCSF get the latent representations of different views in each task, but they consider the relationships of different tasks. On the other hand, compared with MTMVCSF, AN-MTMVCSF introduces noise weight in training stage. We divide each training set into disjoint groups of different sizes. And we compared the performance of PSLF, MTMVCSF and AN-MTMVCSF, and the results are shown in Fig. 5. Obviously, in most data sets, the performance of MTMVCSF and AN-MTMVCSF is better than PSLF. That is to say, considering the relationships of different tasks will improve the performance of the algorithm than the methods that only providing the latent representation of different views. From another aspect, we may find that in most cases MTMVCSF performs better than AN-MTMVCSF, which means in the case of no noise labels the best choice is the algorithm of MTMVCSF which has a better performance and lower complexity. However, when we cannot make sure if there are noise labels, AN-MTMVCSF may be the best choice.

\begin{figure}[!htbp]
	\label{fig5}
	\centering
	\subfigure[Webkb]{
		\includegraphics[width=4.5cm]{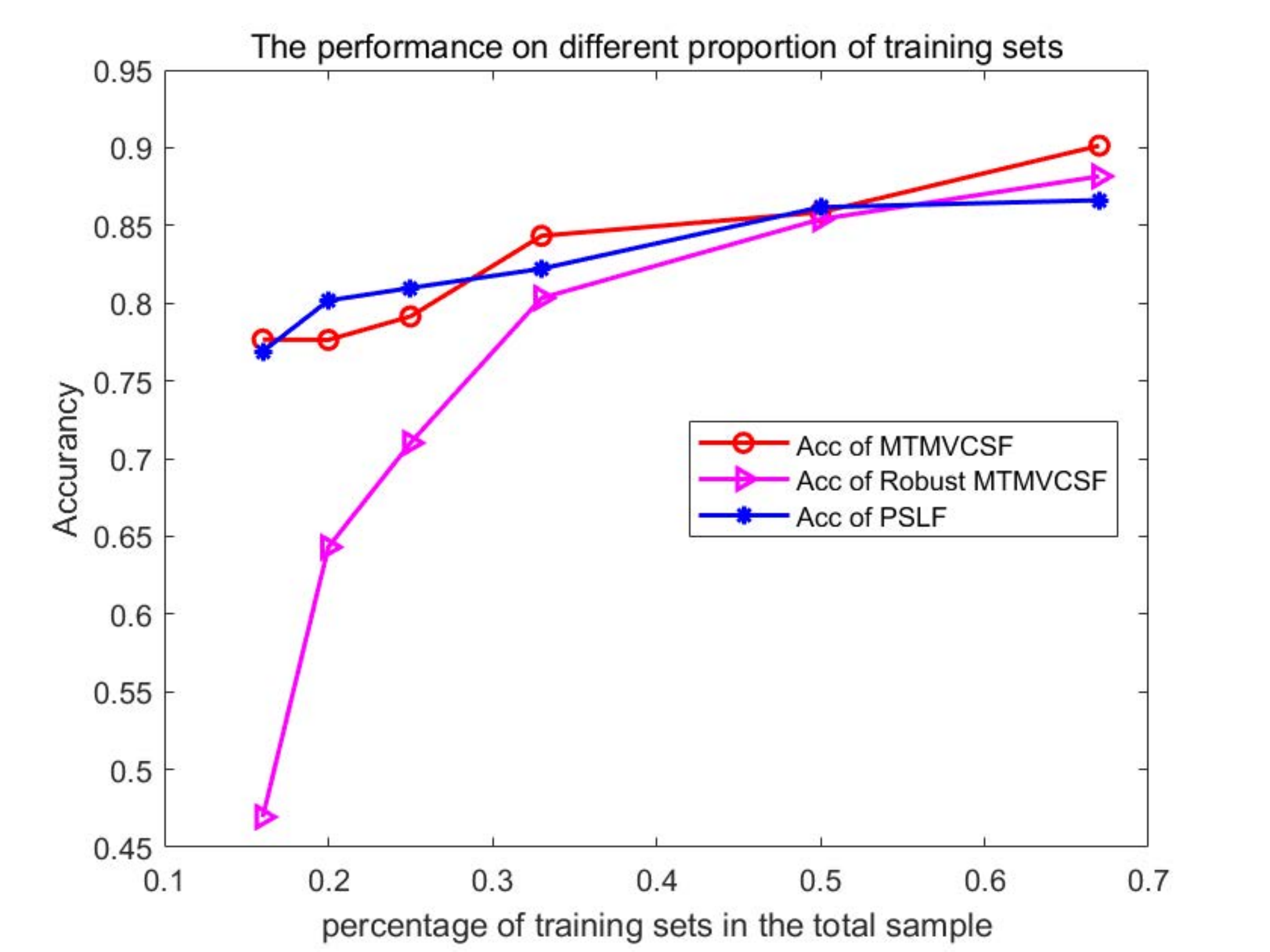}
	}
	\quad
	\subfigure[Leaves]{
		\includegraphics[width=4.5cm]{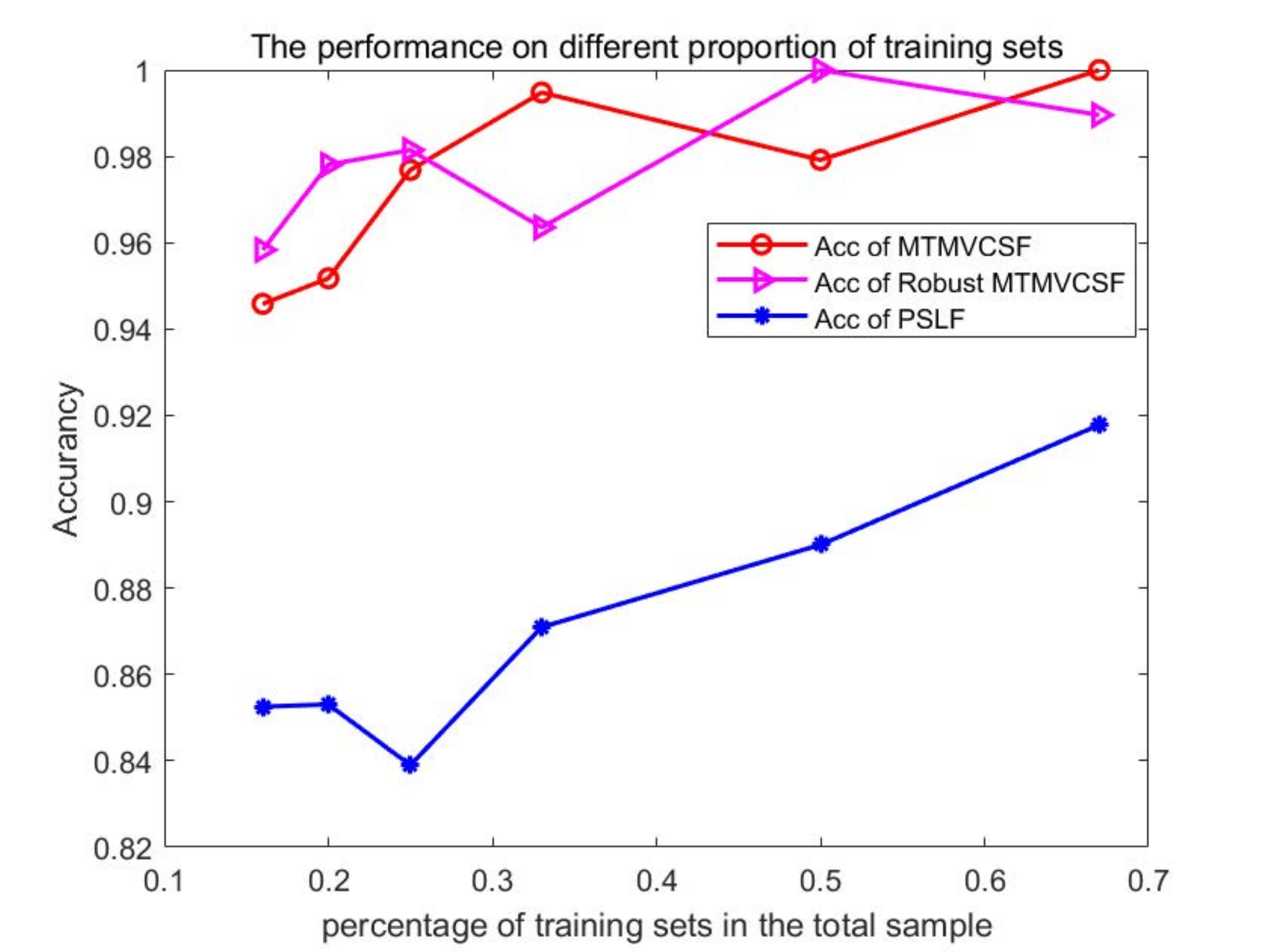}
	}
	\quad
	\subfigure[20NG]{
		\includegraphics[width=4.5cm]{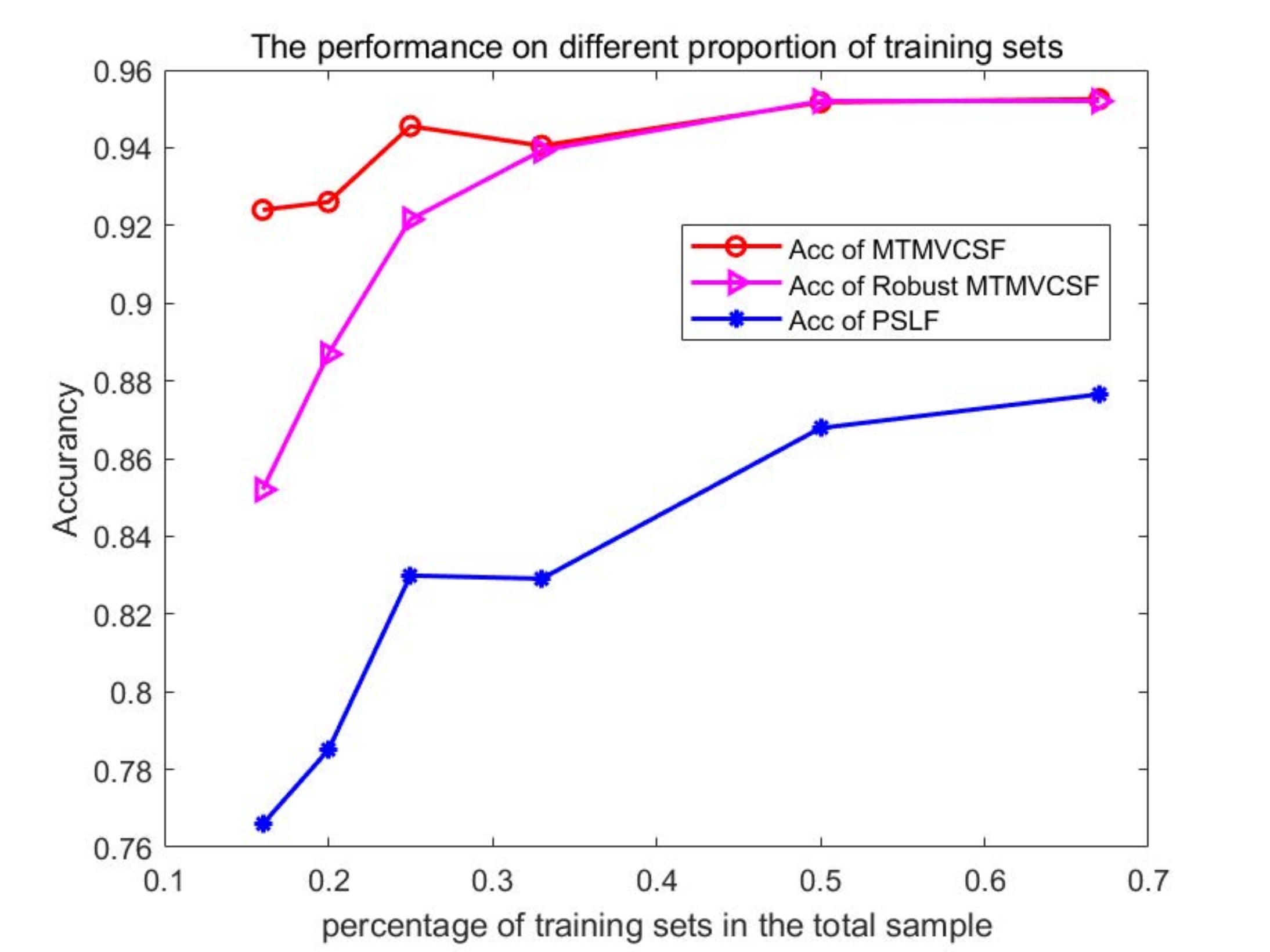}
	}
	\subfigure[NUS-WIDE]{
	     \includegraphics[width=4.5cm]{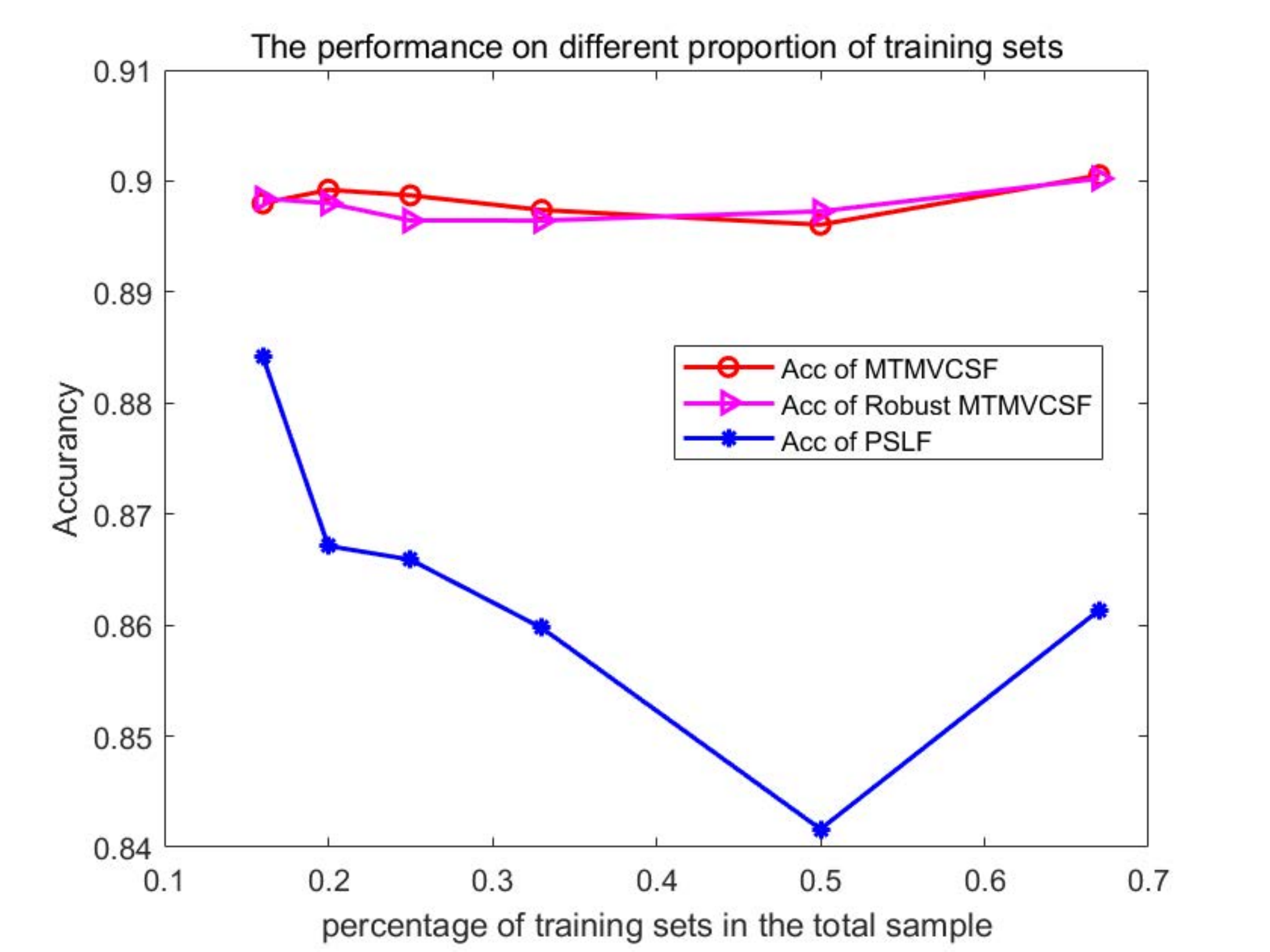}
	}
	\quad
	\subfigure[Synth1]{
		\includegraphics[width=4.5cm]{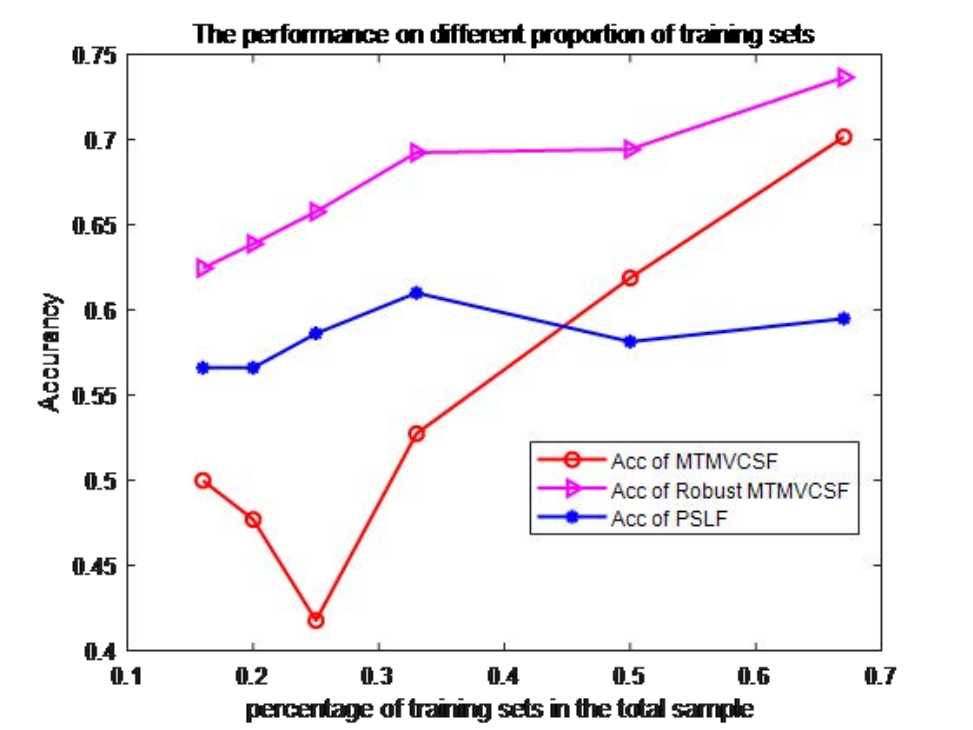}
	}
	\quad
	\subfigure[Synth2]{
		\includegraphics[width=4.5cm]{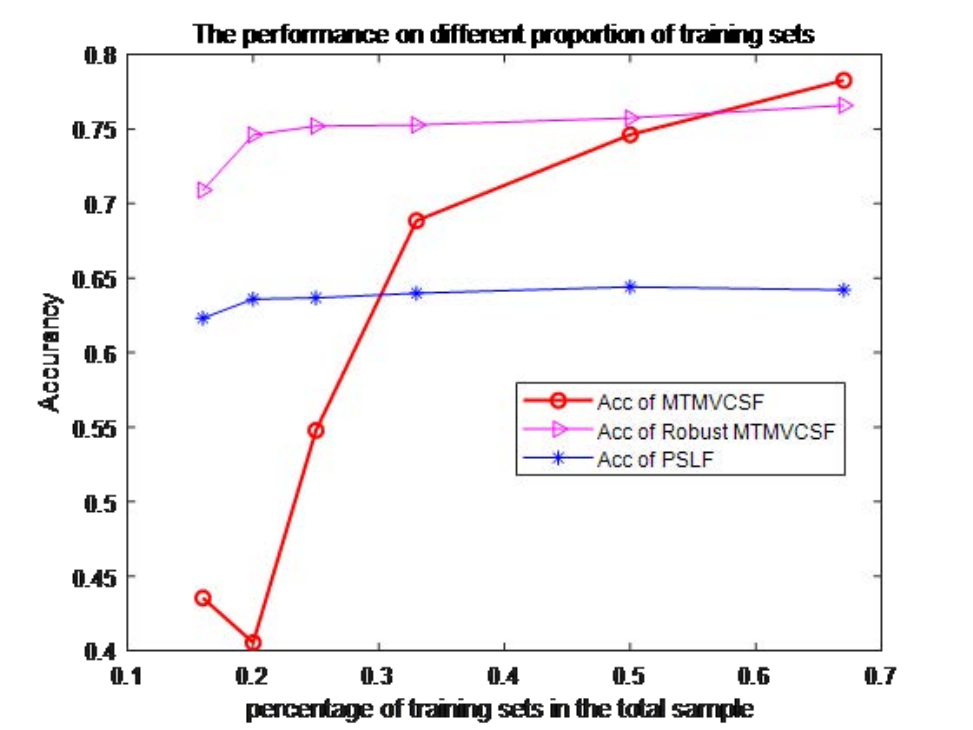}
	}
	\caption{Clustering result in different size of training set with the methods of PSLF and MTMVCSF.}
\end{figure}

\subsection{Conclusion of Experimental Results}

Our proposed methods aim to find the latent features of raw data, and we hope these features can make an improvement on the learning performance. Thus, we want to compare the results of the learning algorithms with different kinds of input data (raw data, preprocessed data using PSLF, preprocessed data using MTMVCSF, and preprocessed data using AN-MTMVCSF). From the results of experiments shown in Table 5 to 8, we actually find that our proposed algorithms can extract good joint features of multiple views and tasks, and on each data, MTMVCSF and AN-MTMVCSF perform better than other selected benchmarks. Besides, from the result shown in Fig. 2, An-MTMVCSF’s anti-noise ability is demonstrated compared with MTMVCSF. And Fig.3, Fig.4, and Table 4 indicate that our proposed methods have well convergence and computing speed. Finally, we prove that the latent features coming from MTMVCSF and AN-MTMVCSF perform better than PSLF according to Fig.5, which also indicates that the relationships among tasks will help the algorithm find better representation.

\section{Conclusion}

Both multi-task and multi-view learning have made significant progress in theory and practice, and joint multi-task and multi-view learning has remained relatively untouched. Aiming at this situation, we provided a novel framework of multi-task multi-view learning via integrating consistent and complementary features of multiple views and tasks. By reconstructing multiple views existing in tasks, we find the latent representation, which both grasp the underlying consistency and complementarity properties of multiple views, each task’s multiple original views are taken placed by each task’s latent representation. Then, we utilized the relationships among tasks to improve algorithm’s learning performance. At the same time, we use fixed point iteration process to derive the model parameters’ updating formulas. Furthermore, we proposed an anti-noise model AN-MTMVCSF to resist noise interference. The experimental results on both real world and synthetic data have verified the effect of our algorithms.

In the future, we plan to use low-rank and sparse method that can find the common and special features among different tasks \cite{18kong2017probabilistic}, and if we use it to place the multi-task method we used, we may consider both common and special information in both view and task level. Another promising research direction is using neural networks to reconstruct views in each task, such as Auto-Encoder \cite{33bengio2009learning}, and this method can improve the nonlinear ability of the reconstruction process. What is more, our proposed method cannot measure the uncertainty of the model, which leads to multiple views’ common factor in each task must be the same. However, multiple views’ common features in each task should come from a same subspace, but their instantiations in multiple views should not be the same. Therefore, a generative model should be proposed to solve this problem, and we can also learn from some popular generative methods, such as variational auto-encoders and generative adversarial nets \cite{46kingma2013auto,47goodfellow2014generative}. Besides, in this paper and some other matrix factorization methods \cite{5liu2014partially,12jin2013shared,25zhang2018multi,34lu2018multi,48li2017lifelong,49lu2017multilinear}, mean square error are used to measure the loss of true value and predict value, however, when we deal with classification or semi-clustering tasks the metric becomes inappropriate, and we should use loss that suitable this kind of problem, such as cross-entropy loss instead of mean square error. Therefore, exploring the supervisory role of different loss functions on constructive implicit representation is also a direction worth studying. Finally, the quality of view construction is the key factor affecting the learning performance, and how to construct good views of raw data is therefore worthy of research. In addition to the methods mentioned in \cite{50xu2013survey}, we can use different feature extraction methods to construct different views and explore their importance in multi-task multi-view learning. For instance, for a document object, we can not only use some traditional method like TF-IDF \cite{51chowdhury2010introduction} and Word2Vec \cite{52mikolov2013efficient}, but can also use some other semantic analysis method like MDLSA \cite{53zhang2013multidimensional}. By comparing the performance of different views, we can find the best combination of views in each task.

\bibliography{bibfile}

\begin{thebibliography}{10}
\expandafter\ifx\csname url\endcsname\relax
  \def\url#1{\texttt{#1}}\fi
\expandafter\ifx\csname urlprefix\endcsname\relax\def\urlprefix{URL }\fi
\expandafter\ifx\csname href\endcsname\relax
  \def\href#1#2{#2} \def\path#1{#1}\fi

\bibitem{1zhang2016multi}
X.~Zhang, X.~Zhang, H.~Liu, X.~Liu, Multi-task multi-view clustering, IEEE
  Transactions on Knowledge and Data Engineering 28~(12) (2016) 3324--3338.

\bibitem{2chaudhuri2009multi}
K.~Chaudhuri, S.~M. Kakade, K.~Livescu, K.~Sridharan, Multi-view clustering via
  canonical correlation analysis, in: Proceedings of the 26th annual
  international conference on machine learning, 2009, pp. 129--136.

\bibitem{3kursun2010canonical}
O.~Kursun, E.~Alpaydin, Canonical correlation analysis for multiview
  semisupervised feature extraction, in: International conference on artificial
  intelligence and soft computing, Springer, 2010, pp. 430--436.

\bibitem{4blum1998combining}
A.~Blum, T.~Mitchell, Combining labeled and unlabeled data with co-training,
  in: Proceedings of the eleventh annual conference on Computational learning
  theory, 1998, pp. 92--100.

\bibitem{5liu2014partially}
J.~Liu, Y.~Jiang, Z.~Li, Z.-H. Zhou, H.~Lu, Partially shared latent factor
  learning with multiview data, IEEE transactions on neural networks and
  learning systems 26~(6) (2014) 1233--1246.

\bibitem{6brefeld2005multi}
U.~Brefeld, C.~B{\"u}scher, T.~Scheffer, Multi-view hidden markov perceptrons.,
  in: LWA, 2005, pp. 134--138.

\bibitem{7zhou2010semi}
Z.-H. Zhou, M.~Li, Semi-supervised learning by disagreement, Knowledge and
  Information Systems 24~(3) (2010) 415--439.

\bibitem{8wang2007analyzing}
W.~Wang, Z.-H. Zhou, Analyzing co-training style algorithms, in: European
  conference on machine learning, Springer, 2007, pp. 454--465.

\bibitem{9he2011graphbased}
J.~He, R.~Lawrence, A graphbased framework for multi-task multi-view learning,
  in: ICML, 2011.

\bibitem{10zhang2012inductive}
J.~Zhang, J.~Huan, Inductive multi-task learning with multiple view data, in:
  Proceedings of the 18th ACM SIGKDD international conference on Knowledge
  discovery and data mining, 2012, pp. 543--551.

\bibitem{11qian2012reconstruction}
B.~Qian, X.~Wang, I.~Davidson, A reconstruction error formulation for
  semi-supervised multi-task and multi-view learning, arXiv preprint
  arXiv:1202.0855 (2012).

\bibitem{12jin2013shared}
X.~Jin, F.~Zhuang, S.~Wang, Q.~He, Z.~Shi, Shared structure learning for
  multiple tasks with multiple views, in: Joint European conference on machine
  learning and knowledge discovery in databases, Springer, 2013, pp. 353--368.

\bibitem{13jin2014multi}
X.~Jin, F.~Zhuang, H.~Xiong, C.~Du, P.~Luo, Q.~He, Multi-task multi-view
  learning for heterogeneous tasks, in: Proceedings of the 23rd ACM
  international conference on conference on information and knowledge
  management, 2014, pp. 441--450.

\bibitem{14hong2013tracking}
Z.~Hong, X.~Mei, D.~Prokhorov, D.~Tao, Tracking via robust multi-task
  multi-view joint sparse representation, in: Proceedings of the IEEE
  international conference on computer vision, 2013, pp. 649--656.

\bibitem{15mei2015robust}
X.~Mei, Z.~Hong, D.~Prokhorov, D.~Tao, Robust multitask multiview tracking in
  videos, IEEE transactions on neural networks and learning systems 26~(11)
  (2015) 2874--2890.

\bibitem{16gong2012robust}
P.~Gong, J.~Ye, C.~Zhang, Robust multi-task feature learning, in: Proceedings
  of the 18th ACM SIGKDD international conference on Knowledge discovery and
  data mining, 2012, pp. 895--903.

\bibitem{17gonccalves2016multi}
A.~R. Gon{\c{c}}alves, F.~J. Von~Zuben, A.~Banerjee, et~al., Multi-task sparse
  structure learning with gaussian copula models, Journal of Machine Learning
  Research (2016).

\bibitem{18kong2017probabilistic}
Y.~Kong, M.~Shao, K.~Li, Y.~Fu, Probabilistic low-rank multitask learning, IEEE
  transactions on neural networks and learning systems 29~(3) (2017) 670--680.

\bibitem{19xiao2019manifold}
L.~Xiao, J.~M. Stephen, T.~W. Wilson, V.~D. Calhoun, Y.-P. Wang, A manifold
  regularized multi-task learning model for iq prediction from multiple fmri
  paradigms, arXiv preprint arXiv:1901.05913 (2019).

\bibitem{20ranjan2017hyperface}
R.~Ranjan, V.~M. Patel, R.~Chellappa, Hyperface: A deep multi-task learning
  framework for face detection, landmark localization, pose estimation, and
  gender recognition, IEEE transactions on pattern analysis and machine
  intelligence 41~(1) (2017) 121--135.

\bibitem{21he2018multi}
Y.~He, J.~Zhang, H.~Shan, L.~Wang, Multi-task gans for view-specific feature
  learning in gait recognition, IEEE Transactions on Information Forensics and
  Security 14~(1) (2018) 102--113.

\bibitem{22zhang2018generalized}
C.~Zhang, H.~Fu, Q.~Hu, X.~Cao, Y.~Xie, D.~Tao, D.~Xu, Generalized latent
  multi-view subspace clustering, IEEE transactions on pattern analysis and
  machine intelligence 42~(1) (2018) 86--99.

\bibitem{23cheng2018tensor}
M.~Cheng, L.~Jing, M.~K. Ng, Tensor-based low-dimensional representation
  learning for multi-view clustering, IEEE Transactions on Image Processing
  28~(5) (2018) 2399--2414.

\bibitem{24yin2018multiview}
Q.~Yin, S.~Wu, L.~Wang, Multiview clustering via unified and view-specific
  embeddings learning, IEEE transactions on neural networks and learning
  systems 29~(11) (2018) 5541--5553.

\bibitem{25zhang2018multi}
Z.~Zhang, Z.~Qin, P.~Li, Q.~Yang, J.~Shao, Multi-view discriminative learning
  via joint non-negative matrix factorization, in: International Conference on
  Database Systems for Advanced Applications, Springer, 2018, pp. 542--557.

\bibitem{26sun2013multi}
S.~Sun, G.~Chao, Multi-view maximum entropy discrimination, in: Twenty-third
  international joint conference on artificial intelligence, 2013.

\bibitem{27chao2015alternative}
G.~Chao, S.~Sun, Alternative multiview maximum entropy discrimination, IEEE
  transactions on neural networks and learning systems 27~(7) (2015)
  1445--1456.

\bibitem{28fang2014detecting}
Y.~Fang, H.~Zhang, Y.~Ye, X.~Li, Detecting hot topics from twitter: A multiview
  approach, Journal of Information Science 40~(5) (2014) 578--593.

\bibitem{29cui2018mv}
Q.~Cui, S.~Wu, Q.~Liu, W.~Zhong, L.~Wang, Mv-rnn: A multi-view recurrent neural
  network for sequential recommendation, IEEE Transactions on Knowledge and
  Data Engineering 32~(2) (2018) 317--331.

\bibitem{30cui2018mv}
Q.~Cui, S.~Wu, Q.~Liu, W.~Zhong, L.~Wang, Mv-rnn: A multi-view recurrent neural
  network for sequential recommendation, IEEE Transactions on Knowledge and
  Data Engineering 32~(2) (2018) 317--331.

\bibitem{31ma2018learning}
C.~Ma, Y.~Guo, J.~Yang, W.~An, Learning multi-view representation with lstm for
  3-d shape recognition and retrieval, IEEE Transactions on Multimedia 21~(5)
  (2018) 1169--1182.

\bibitem{32huang2018deep}
F.~Huang, X.~Zhang, Z.~Zhao, Z.~Li, Y.~He, Deep multi-view representation
  learning for social images, Applied Soft Computing 73 (2018) 106--118.

\bibitem{33bengio2009learning}
Y.~Bengio, Learning deep architectures for AI, Now Publishers Inc, 2009.

\bibitem{34lu2018multi}
R.~Lu, X.~Zuo, J.~Liu, S.~Lian, Multi-view learning based on common and special
  features in multi-task scenarios, in: 2018 37th Chinese control conference
  (CCC), IEEE, 2018, pp. 9410--9415.

\bibitem{35wu2018dmtmv}
Y.-F. Wu, D.-C. Zhan, Y.~Jiang, Dmtmv: a unified learning framework for deep
  multi-task multi-view learning, in: 2018 IEEE international conference on big
  knowledge (ICBK), IEEE, 2018, pp. 49--56.

\bibitem{36sun2018robust}
G.~Sun, Y.~Cong, J.~Li, Y.~Fu, Robust lifelong multi-task multi-view
  representation learning, in: 2018 IEEE international conference on big
  knowledge (ICBK), IEEE, 2018, pp. 91--98.

\bibitem{37zhang2015mmfe}
Q.~Zhang, L.~Zhang, B.~Du, W.~Zheng, W.~Bian, D.~Tao, Mmfe: Multitask multiview
  feature embedding, in: 2015 IEEE International Conference on Data Mining,
  IEEE, 2015, pp. 1105--1110.

\bibitem{38yan2013no}
Y.~Yan, E.~Ricci, R.~Subramanian, O.~Lanz, N.~Sebe, No matter where you are:
  Flexible graph-guided multi-task learning for multi-view head pose
  classification under target motion, in: Proceedings of the IEEE international
  conference on computer vision, 2013, pp. 1177--1184.

\bibitem{39yan2014evaluating}
Y.~Yan, R.~Subramanian, E.~Ricci, O.~Lanz, N.~Sebe, Evaluating multi-task
  learning for multi-view head-pose classification in interactive environments,
  in: 2014 22nd international conference on pattern recognition, IEEE, 2014,
  pp. 4182--4187.

\bibitem{40kandemir2014multi}
M.~Kandemir, A.~Vetek, M.~Goenen, A.~Klami, S.~Kaski, Multi-task and multi-view
  learning of user state, Neurocomputing 139 (2014) 97--106.

\bibitem{41javanmardi2018robust}
M.~Javanmardi, X.~Qi, Robust structured multi-task multi-view sparse tracking,
  in: 2018 IEEE international conference on multimedia and expo (ICME), IEEE,
  2018, pp. 1--6.

\bibitem{42han2017multi}
H.~Han, S.~Kang, C.~D. Yoo, Multi-view visual speech recognition based on multi
  task learning, in: 2017 IEEE international conference on image processing
  (ICIP), IEEE, 2017, pp. 3983--3987.

\bibitem{43li2017fast}
Q.~Li, Z.~Sun, R.~He, Fast multi-view face alignment via multi-task
  auto-encoders, in: 2017 IEEE international joint conference on biometrics
  (IJCB), IEEE, 2017, pp. 538--545.

\bibitem{44evgeniou2007multi}
A.~Evgeniou, M.~Pontil, Multi-task feature learning, Advances in neural
  information processing systems 19 (2007) 41.

\bibitem{45wold1987principal}
S.~Wold, K.~Esbensen, P.~Geladi, Principal component analysis, Chemometrics and
  intelligent laboratory systems 2~(1-3) (1987) 37--52.

\bibitem{46kingma2013auto}
D.~P. Kingma, M.~Welling, Auto-encoding variational bayes, arXiv preprint
  arXiv:1312.6114 (2013).

\bibitem{47goodfellow2014generative}
I.~Goodfellow, J.~Pouget-Abadie, M.~Mirza, B.~Xu, D.~Warde-Farley, S.~Ozair,
  A.~Courville, Y.~Bengio, Generative adversarial nets, Advances in neural
  information processing systems 27 (2014).

\bibitem{48li2017lifelong}
X.~Li, S.~N. Chandrasekaran, J.~Huan, Lifelong multi-task multi-view learning
  using latent spaces, in: 2017 IEEE international conference on big data (Big
  Data), IEEE, 2017, pp. 37--46.

\bibitem{49lu2017multilinear}
C.-T. Lu, L.~He, W.~Shao, B.~Cao, P.~S. Yu, Multilinear factorization machines
  for multi-task multi-view learning, in: Proceedings of the tenth ACM
  international conference on web search and data mining, 2017, pp. 701--709.

\bibitem{50xu2013survey}
C.~Xu, D.~Tao, C.~Xu, A survey on multi-view learning, arXiv preprint
  arXiv:1304.5634 (2013).

\bibitem{51chowdhury2010introduction}
G.~G. Chowdhury, Introduction to modern information retrieval, Facet
  publishing, 2010.

\bibitem{52mikolov2013efficient}
T.~Mikolov, K.~Chen, G.~Corrado, J.~Dean, Efficient estimation of word
  representations in vector space, arXiv preprint arXiv:1301.3781 (2013).

\bibitem{53zhang2013multidimensional}
H.~Zhang, J.~K. Ho, Q.~J. Wu, Y.~Ye, Multidimensional latent semantic analysis
  using term spatial information, IEEE transactions on cybernetics 43~(6)
  (2013) 1625--1640.

\end{thebibliography}

\end{document}